%% file: main.tex
\documentclass{article}

\usepackage{mafm_preprint,times}
\input{math_commands.tex}

\usepackage{hyperref}
\usepackage{url}
\usepackage{booktabs}
\usepackage{tabularx}
\usepackage{multirow}
\usepackage{graphicx}
\usepackage{subcaption}
\usepackage{amsmath,amssymb,amsthm}
\usepackage{mathtools}
\usepackage{xcolor}
\colorlet{revisionink}{black}
\colorlet{clarityink}{black}
\colorlet{costink}{black}
\DeclareRobustCommand{\rev}[1]{{\color{revisionink}#1}}
\DeclareRobustCommand{\clarrev}[1]{{\color{clarityink}#1}}
\DeclareRobustCommand{\costrev}[1]{{\color{costink}#1}}
\newenvironment{revision}[1][revisionink]{\begingroup\color{#1}}{\par\endgroup}
\usepackage{microtype}
\usepackage{enumitem}
\usepackage{algorithm}
\usepackage{algpseudocode}
\usepackage{placeins}
\usepackage{float}

\hypersetup{hidelinks}
\renewcommand{\eqref}[1]{(\ref{#1})}

\newtheorem{proposition}{Proposition}
\newtheorem{corollary}[proposition]{Corollary}

\newcommand{\C}{\mathcal C}
\newcommand{\Sdata}{\mathcal S}
\newcommand{\Ncon}{\mathcal N}
\newcommand{\M}{\mathcal M}
\newcommand{\dd}{\mathrm d}
\newcommand{\method}{MAFM}
\newcommand{\methodfull}{Manifold Attracting Flow Matching}
\newcommand{\feas}{\textsc{Y}}
\newcommand{\infeas}{\textsc{N}}
\newcommand{\best}[1]{\textbf{#1}}
\newcommand{\second}[1]{\underline{#1}}

\newcommand{\papertitle}{Manifold Attracting Flow Matching: Physics-Constrained Sampling via Continuous Geometric Dynamics}
\title{\papertitle}
\hypersetup{
  pdftitle={\papertitle},
  pdfauthor={Guanyu Chen, Jinchen Wang, Shengze Xu, Xingyu Ren, Tieyong Zeng, Antonios Armaou, Dong Ni}
}

\newcommand{\authorblock}{%
  {\large\bfseries
    Guanyu Chen\textsuperscript{1,*}, Jinchen Wang\textsuperscript{2,*},
    Shengze Xu\textsuperscript{3}, Xingyu Ren\textsuperscript{1}\\
    Tieyong Zeng\textsuperscript{3}, Antonios Armaou\textsuperscript{4},
    Dong Ni\textsuperscript{1,\ensuremath{\dagger}}\par}
  \vspace{0.65em}
  {\small
    \textsuperscript{1}Zhejiang University\qquad
    \textsuperscript{2}University of California, Los Angeles\\
    \textsuperscript{3}The Chinese University of Hong Kong\qquad
    \textsuperscript{4}Pennsylvania State University\par}
}

\author{\authorblock}

\begin{document}
\maketitle
\begingroup
  \renewcommand{\thefootnote}{\fnsymbol{footnote}}
  \footnotetext[1]{Equal contribution.}
  \footnotetext[2]{Corresponding author:
    \href{mailto:dni@zju.edu.cn}{\texttt{dni@zju.edu.cn}}.}
\endgroup

\begin{abstract}
Generating scientific data under previously unseen equality constraints requires satisfying the prescribed constraints without destroying the physical structure learned from data. Existing inference-time approaches often enforce physical constraints by repeatedly correcting predicted endpoints, but such discrete corrections can distort learned generative structure and introduce unstable feedback. In this work, we propose Manifold Attracting Flow Matching (MAFM), which continuously evolves a persistent anchor through an ODE whose dynamics drive it analytically toward the constraint manifold. Specifically, its geometric correction preserves model-guided tangent motion while prescribing the normal motion to contract constraint violations toward the manifold. This construction separates model tracking from constraint attraction and provides theoretical guarantees for residual contraction and local feedback stabilization. Experiments across geometric and partial differential equation benchmarks demonstrate improved generation quality over variant constraint-guided baselines while maintaining low constraint residuals. MAFM provides a flexible framework for coordinating learned physical structure and explicit constraints through controllable sampling dynamics, without retraining the underlying generative model.
\end{abstract}

\input{sections/introduction}

\input{sections/related_work}
\input{sections/method}
\input{sections/experiments}
\input{sections/conclusion}

\clearpage
\input{sections/statements}

\bibliography{references}
\bibliographystyle{iclr2027_conference}
\newpage
\appendix
\input{sections/appendix}

\end{document}

%% file: math_commands.tex
\usepackage{amsmath,amsfonts,bm}

\def\eqref#1{equation~\ref{#1}}

\def\1{\bm{1}}

\DeclareMathAlphabet{\mathsfit}{\encodingdefault}{\sfdefault}{m}{sl}
\SetMathAlphabet{\mathsfit}{bold}{\encodingdefault}{\sfdefault}{bx}{n}



%% file: sections/introduction.tex
\section{Introduction}
\label{sec:introduction}



Generative models have become an increasingly important tool for producing diverse and realistic samples across scientific and engineering domains. Yet in many applications, realism alone is not sufficient: we need to generate samples that satisfy what the task actually asks for. Crucially, many of these requirements are \emph{hard} rather than preferential---a sample that violates them may be plausible under the learned data distribution, but still invalid for downstream use. For example, PDE reconstruction from sparse observations requires generated solution fields to agree with prescribed measurements or boundary values \citep{huang2024diffusionpde, shysheya2024conditional}; controllable image generation may require specified regions or pixels to obey a prescribed spatial or color pattern \citep{endo2024masked, shum2025color}; and safety-critical decision making requires states or actions to remain within prescribed safety limits rather than merely reducing violations on average \citep{Ames_2017, shi2023near, hu2025uncertain}. These examples highlight a common need: generating diverse samples while satisfying specific constraints exactly.

A particularly important setting arises when such constraints are specified only after training. A generative model may be trained once on a broad data distribution, but subsequently queried with new observations, boundary values, geometric specifications, or control targets \citep{wei2024diffphycon, feng2025guidance}. Collecting a dedicated dataset or retraining a conditional model for every new request would substantially limit the reuse of the learned generative prior. We therefore study \emph{inference-time constrained generation}: an unconditional model learns the data distribution independently of downstream constraints, while an analytic equality constraint is supplied only to the sampler at inference time. \rev{Our goal is to generate a diverse set of feasible samples that obey the governing physical laws and, for PDE tasks, lie on the solution manifold learned by the model.}

Existing approaches incorporate physical information at different stages.
Physics-informed training and constrained fine-tuning modify the learning
objective to account for known physical requirements
\citep{bastek2025physics,tauberschmidt2026physics}.  Inference-time methods instead
reuse a pretrained model: DiffusionPDE guides generation using residual
gradients \citep{huang2024diffusionpde}, while ECI alternates endpoint
extrapolation, physical correction, and interpolation
\citep{cheng2025gradient}.  Physics-Constrained Flow Matching (PCFM)
corrects predicted flow endpoints and re-interpolates with the original
source to enforce sampling-time constraints \citep{utkarsh2026physics}.
Despite these advances, satisfying the prescribed constraint is only one
requirement: constraint enforcement should not drive generated samples away from
the manifold learned from data. We therefore ask whether a sampler can satisfy a
prescribed hard constraint while preserving fidelity to the learned data manifold.
Accordingly, we evaluate two complementary objectives: constraint satisfaction
and data-manifold fidelity. \rev{We report distributional statistics as
supplementary comparisons against specified reference ensembles.}

We introduce \methodfull{} (\method), an inference-time method for imposing
new equality constraints on a pretrained generative model without retraining.
The key idea is \emph{persistent-anchor dynamics}. Instead of injecting constraint
guidance directly into the transient sampling state $x_t$, as in gradient-guidance
or projection-based methods, \method{} guides constrained generation through the
continuous geometric evolution of a persistent anchor, whose dynamics analytically
converge to the constraint manifold. Learned endpoint predictions and constraint
feedback are jointly accumulated in this anchor, which serves as the evolving
candidate solution ultimately returned by the method. Its dynamics separate the
assimilation of constraint-compatible model information from attraction toward the
constraint manifold, allowing learned structure to be preserved while feasibility
is controlled at an independently prescribed rate. We characterize the resulting
constraint-residual dynamics under explicit assumptions and analyze how
endpoint-prediction and adjoint-estimation errors propagate through the anchor
trajectory.

We evaluate \method{} across five scientific-computing benchmarks involving
Poisson, Burgers, \mbox{reaction--diffusion} (RD), Euler, and Navier--Stokes (NS)
systems. Together, these tasks test constraints arising from partial
observations, conservation laws, prescribed physical quantities, and recovery
of feasible samples while retaining learned data structure. \rev{On all five
PDE benchmarks reported in Table~\ref{tab:dynamics-main}, \method{} meets the
prescribed constraint tolerance and achieves lower sample-level relative $L_2$
error than the baselines that meet the same tolerance.}
So our contributions can be summarized as follows:
\begin{itemize}[leftmargin=*,itemsep=2pt,topsep=3pt]

\item We introduce a \emph{persistent-anchor} sampling paradigm for imposing new
equality constraints at inference time with a frozen flow model. The persistent
anchor provides a shared endpoint representation that continuously integrates
learned guidance from the pretrained flow with constraint feedback, directing
both toward the eventual output.

\item We characterize the constraint-residual dynamics induced by the
persistent-anchor updates, establish conditions for convergence toward
feasibility, and analyze how endpoint-prediction and adjoint-estimation errors
affect the anchor trajectory.

\item We evaluate \method{} across scientific-computing benchmarks involving
partial observations, conservation laws, and prescribed physical quantities,
demonstrating that hard constraints can be satisfied to numerical tolerance
while preserving fidelity to the learned data support.
\end{itemize}

%% file: sections/related_work.tex
\section{Related Work}
\label{sec:related}

\paragraph{Generative modeling for physical systems.}
Continuous normalizing flows and flow matching provide flexible generative models for high-dimensional physical states \citep{chen2018neural,papamakarios2021normalizing,lipman2022flow}, with minibatch optimal transport often used to simplify conditional transport paths \citep{tong2023improving}. When the underlying geometry is known, Riemannian score models and flow matching can construct dynamics directly on the prescribed manifold \citep{de2022riemannian,chen2024flow}. In contrast, our setting learns the data support from ambient samples and introduces a separate analytic constraint only at inference time.

Physical information has been incorporated either during training or during sampling. Residual-based objectives and physics-constrained fine-tuning bias the learned model toward known physical requirements \citep{raissi2019physics,bastek2025physics,jacobsen2025cocogen,baldan2026physics,tauberschmidt2026physics}. Inference-time approaches instead reuse a pretrained model: DiffusionPDE applies residual-gradient guidance \citep{huang2024diffusionpde}, ECI alternates endpoint extrapolation with hard correction and interpolation \citep{cheng2025gradient}, and PCFM repeatedly corrects predicted endpoints before re-interpolation \citep{utkarsh2026physics}. \rev{PnP-Flow~\citep{martin2025pnpflowplugandplayimagerestoration} also uses reinterpolation, alternating data fidelity and denoising for inverse problems. Lagrangian Dual Flows~\citep{kurtz2026constrainedflowmatchinglagrangian} evolves a constraint dual variable with the sample. D-Flow~\citep{benhamu2024dflowdifferentiatingflowscontrolled} differentiates through the full generation ODE to optimize the source noise. FunDPS~\citep{yao2026guideddiffusionsamplingfunction} uses gradients through a denoised estimate in function-space posterior sampling, a guidance operation tested on our FM backbone by the full-gradient baseline.} However, constraint satisfaction alone does not ensure that these corrections preserve the physical relationships learned from data. While operating at inference time with a frozen flow model, \method{} maintains a \emph{persistent endpoint anchor} across sampling steps, using learned endpoint guidance and constraint feedback to update a common candidate for the final output rather than attaching constraint control only to transient sampling states.

\paragraph{Geometric motivation and constraint stabilization.}
Stiff and fast--slow dynamical systems naturally separate rapid attraction toward a lower-dimensional set from slower evolution along it \citep{wanner1996solving,FENICHEL197953}. Classical constrained integration similarly uses algebraic multipliers to enforce or stabilize equality constraints \citep{gear1988differential,ascher1993stability}, with related ideas appearing in neural differential-algebraic models \citep{boesen2025neural,diamond2025neural}. These perspectives motivate the separation used in \method{}: learned endpoint information primarily guides motion compatible with the feasible geometry, while multiplier-based feedback independently controls contraction of the known constraint residual. We use this viewpoint as algorithmic motivation; our analysis establishes residual dynamics for the prescribed constraint and does not assume persistence of the unknown learned data support.

%% file: sections/method.tex
\section{Manifold Attracting Flow Matching}
\label{sec:method}

\begin{figure}[t]
  \centering
  \includegraphics[width=\linewidth,trim=35bp 207bp 35bp 205bp,clip]{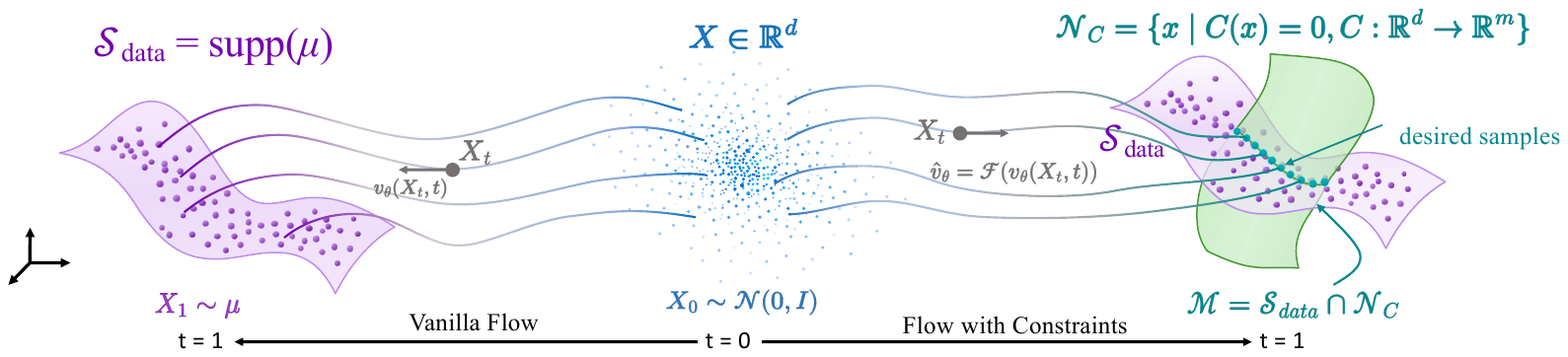}
  \caption{\textbf{Constrained sampling on a learned data support.}
  The broad data support (purple) meets the supplied constraint set (green)
  at the desired feasible intersection.  The illustration denotes the data
  distribution by $\mu=p_{\mathrm{data}}$, its support by
  $\mathcal S_{\mathrm{data}}=\Sdata$, and the constraint set by
  $\mathcal N_{\C}=\Ncon$.}
  \label{fig:setting}
\end{figure}

We first formulate inference-time constrained generation as sampling where a
learned data support meets a new equality constraint. We then derive
persistent-anchor dynamics that combine frozen-model endpoint guidance with
multiplier-based constraint control. Finally, we give a discrete sampler and
analyze residual contraction, adjoint guidance, and finite-step error.

\subsection{Problem formulation}

Let $\mathcal X=\mathbb R^d$ be the space of complete physical objects.  We
write the source and data endpoints as
$x_0\sim p_0=\mathcal N(0,I_d), x_1\sim p_{\mathrm{data}},$
and denote the data support by $\Sdata=\operatorname{supp}(p_{\mathrm{data}})$.
At generation time, an equality constraint specifies
\begin{equation}
  \C:\mathcal X\rightarrow\mathbb R^m,
  \qquad \Ncon=\{x\in\mathcal X:\C(x)=0\},
  \label{eq:constraint-set}
\end{equation}
The feasible intersection is $\M=\Sdata\cap\Ncon \neq\varnothing $ as is shown in Figure~\ref{fig:setting}.  We use
$\mu_q$ for a desired control distribution whose support is
contained in $\M$, without prescribing probability weights on the
intersection.  The frozen unconditional model learns relationships in the
broad data distribution; test-time constraints enter only the sampler.
This assumption places the requested constraint within the physical range
represented by the data.  It allows the finite training set to contain no
exactly feasible sample.  We assess the generated output through two
complementary objectives,
\begin{equation}
  \mathfrak e_{\mathrm{con}}(x)=\|\C(x)\|,\qquad
  \mathfrak e_{\mathrm{supp}}(x)=\operatorname{dist}(x,\Sdata).
  \label{eq:two-objectives}
\end{equation}
The unknown support makes the second objective difficult to compute directly.
Geometry, PDE residuals, and independent rollouts test specified relationships
in the data; distribution statistics provide auxiliary coverage diagnostics.
An additional choice of probability weights and its corresponding evaluation
appear in Appendix~\ref{app:target-geometry}.

\paragraph{Global and local constraints.}
For a differentiable constraint, define the row-oriented Jacobian
$J_{\C}(a)=\partial\C(a)/\partial a\in\mathbb R^{m\times d}$.
The product $J_{\C}(a)u$ gives the residual rate induced by a physical
velocity $u$, while $J_{\C}(a)^\top\lambda$ gives a physical-space normal
correction. \clarrev{Here $\lambda\in\mathbb R^m$ contains one Lagrange multiplier per scalar constraint.}  A \emph{global constraint} has a correction distributed broadly
over the physical object, as for mass or energy.  A \emph{local observation
constraint} has a correction supported at selected sensors, windows, or
channels, as in $\C(x)=Hx-y$.  \clarrev{Even a single integral constraint can induce a correction across the
entire field. For sparse observations, we use optional input guidance through
the frozen model to carry constraint information to unobserved coordinates.}

\paragraph{Frozen endpoint model.}
Let $(x_0,x_1)\sim\pi$, with marginals $p_0$ and $p_{\mathrm{data}}$;
the coupling $\pi$ need not be independent.  Vanilla flow matching uses the
complete-space training interpolation
\begin{equation}
  x_t^{\mathrm{train}}=(1-t)x_0+t x_1,\qquad t\in[0,1],
  \label{eq:linear-path}
\end{equation}
whose target velocity is $x_1-x_0$.  The frozen model defines the Vanilla
flow $\dot x_t^{\mathrm{FM}}=v_\theta(x_t^{\mathrm{FM}},t)$ from
$x_{t=0}^{\mathrm{FM}}=x_0$ \citep{lipman2022flow}.  We extract endpoint
information through
\begin{equation}
  D_\theta(x,t)=x+(1-t)v_\theta(x,t),
  \label{eq:endpoint-estimator}
\end{equation}
an endpoint-extrapolation form also used in PCFM \citep{utkarsh2026physics}.
Under finite second moments, for the population-optimal regression field
$v^*(x,t)=\mathbb E[x_1-x_0\mid x_t^{\mathrm{train}}=x]$,
\begin{equation}
  D^*(x,t)=x+(1-t)v^*(x,t)
          =\mathbb E[x_1\mid x_t^{\mathrm{train}}=x]
  \label{eq:endpoint-mean}
\end{equation}
on the training path almost everywhere.  Thus $D_\theta$ is a learned
data-endpoint estimate used repeatedly by the anchor dynamics.  Its proof,
parameterization, and query-region conditions are given in
Appendix~\ref{app:endpoint-details}.

\begin{figure}[t]
  \centering
  \includegraphics[width=\linewidth]{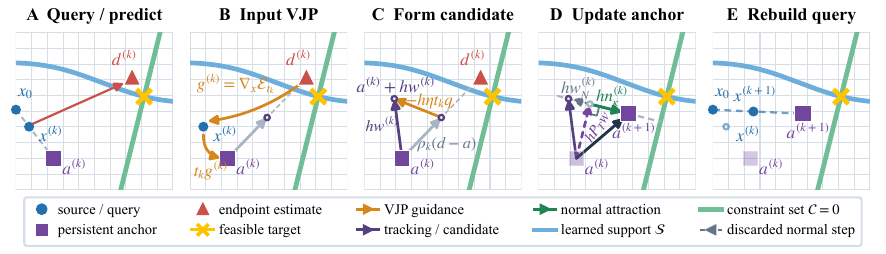}
  \caption{\clarrev{One MAFM sampling step, illustrated with an affine constraint and an exact, unregularized solve. (A) Predict the physical endpoint. (B) Compute optional input guidance. (C) Combine it with endpoint tracking to form the candidate velocity. (D) The anchor update retains the candidate's tangential component and replaces its normal component to shrink the anchor residual. (E) The fixed source and updated anchor give the next model query $x^{(k+1)}=(1-t_{k+1})x_0+t_{k+1}a^{(k+1)}$. At $t=1$, the query equals the final anchor.}}
  \label{fig:pipeline}
\end{figure}
\subsection{Guided Manifold Attracting Dynamics}
We construct the sampling dynamics from a persistent endpoint anchor, optional
guidance through the frozen model, and a multiplier that controls the anchor's
constraint residual.
\paragraph{Query and persistent anchor.}
\clarrev{The new constraint concerns the final physical object.}  For each fixed source
$x_0$, MAFM maintains a network query $x_t$ and a persistent endpoint anchor
$a_t$, both in $\mathcal X$, with
\begin{equation}
  x_t=(1-t)x_0+t a_t.
  \label{eq:query-state}
\end{equation}
The query starts at $x_{t=0}=x_0$ and is rebuilt after every anchor update.
At $t=1$ the two states coincide, so the generated output is
$x_1^{\mathrm{gen}}:=x_{t=1}=a_1$.  Here $x_1$ without a superscript remains
the data endpoint in Eq.~\eqref{eq:linear-path}.  \clarrev{Repeated predictions from
the frozen endpoint model guide this anchor, while the correction below
controls its constraint residual.}

\paragraph{Endpoint-guided candidate.}
For local observations, direct normal correction acts on the measured
coordinates.  The frozen endpoint model can also carry their mismatch to
other coordinates through its input sensitivity.  To obtain this additional
guidance, define
\begin{equation}
  \mathcal E_t(x)=\frac12\|\C(D_\theta(x,t))\|^2,
  \label{eq:endpoint-constraint-energy}
\end{equation}
This energy measures the constraint mismatch
of the predicted endpoint; the normal correction below acts on the anchor
residual $\C(a_t)$.  When this option is active, the input adjoint at $x_t$ is
\begin{equation}
  g_t=\nabla_x\mathcal E_t(x_t)
  =J_{D_\theta}(x_t,t)^\top
   J_{\C}(D_\theta(x_t,t))^\top
   \C(D_\theta(x_t,t)).
  \label{eq:input-adjoint}
\end{equation}
The query relation in Eq.~\eqref{eq:query-state} has anchor derivative
$tI_d$.  Its chain rule therefore gives
\begin{equation}
  \left.\nabla_a\mathcal E_t((1-t)x_0+ta)\right|_{a=a_t}=t\,g_t.
  \label{eq:anchor-adjoint}
\end{equation}
The Jacobian transposes transfer the predicted endpoint's residual to the
complete query; Eq.~\eqref{eq:anchor-adjoint} then transfers it to the anchor.
The model parameters remain frozen. \clarrev{Here $J_{D_\theta}(x_t,t)$ is the Jacobian of the endpoint predictor
with respect to $x_t$ at fixed $t$. In Eq.~\eqref{eq:input-adjoint}, we use a
vector--Jacobian product (VJP), a standard operation in reverse-mode automatic
differentiation, to apply $J_{D_\theta}(x_t,t)^\top$ without explicitly forming
the model Jacobian.}

Let $\gamma(t),\eta(t)\geq0$ be the tracking and guidance rates.
For $\gamma(t)>0$, the optional adjoint can
be expressed as an endpoint displacement.  Set
$\beta(t)=\eta(t)/\gamma(t)$ and define
\begin{equation}
  \widetilde D_\theta(x_t,t)=D_\theta(x_t,t)-\beta(t)t g_t.
  \label{eq:virtual-endpoint}
\end{equation}
The factor $t$ follows from Eq.~\eqref{eq:anchor-adjoint}.  For all
nonnegative rates, the candidate velocity is defined by
\begin{equation}
  w_t=
  \underbrace{\gamma(t)[D_\theta(x_t,t)-a_t]}_{\text{endpoint tracking}}
  -
  \underbrace{\eta(t)t g_t}_{\text{endpoint-constraint guidance}}.
  \label{eq:guided-candidate}
\end{equation}
When $\gamma(t)>0$, this equals
$\gamma(t)[\widetilde D_\theta(x_t,t)-a_t]$; the virtual endpoint is an
equivalent interpretation, not an additional state.  At $\gamma(t)=0$,
Eq.~\eqref{eq:guided-candidate} still gives $w_t=-\eta(t)t g_t$.
\rev{Setting $\eta\equiv0$ skips the input VJP and retains finite-rate
endpoint tracking. For a full-row-rank affine constraint and an exact,
unregularized solve, we recover the PCFM projection and reinterpolation step
when $\eta=0$, the tangential tracking weight reaches its full-replacement
limit $\rho_k=1$, and the normal correction satisfies $h\kappa_k=1$.
Appendix~\ref{app:anchor-memory} gives the two recurrences.} The Poisson experiment uses a fixed channel-scaled
adjoint direction, specified in Appendix~\ref{app:poisson-scaling}; the
multiplier construction is unchanged.

\paragraph{Constraint-compatible anchor dynamics.}
Let $G(a)=J_{\C}(a)J_{\C}(a)^\top$.  An algebraic multiplier corrects the
combined candidate velocity at the anchor, giving the unified dynamics
\begin{align}
  (G(a_t)+\delta I_m)\lambda_t
    &=-\left[J_{\C}(a_t)w_t+\kappa(t)\C(a_t)\right],
    \label{eq:unified-multiplier}\\
  \dot a_t&=w_t+J_{\C}(a_t)^\top\lambda_t.
  \label{eq:unified-anchor-ode}
\end{align}
\costrev{The rate $\kappa(t)\geq0$ controls how quickly the anchor residual decays. Its time integral sets the cumulative contraction.}
$\delta\geq0$ regularizes the Gram solve.  For full-row-rank $J_{\C}$, define
$P_T(a)=I_d-J_{\C}(a)^\top G(a)^{-1}J_{\C}(a)$.  With an exact
unregularized solve ($\delta=0$), the dynamics become
\begin{equation}
  \boxed{
  \dot a_t=P_T(a_t)w_t
  -J_{\C}(a_t)^\top G(a_t)^{-1}\kappa(t)\C(a_t)}.
  \label{eq:unified-tangent-normal}
\end{equation}
The first term retains the candidate's constraint-neutral component:
\begin{equation*}
  P_T(a_t)w_t=\gamma(t)P_T(a_t)[D_\theta(x_t,t)-a_t]-\eta(t)tP_T(a_t)g_t.
\end{equation*}
Thus the adjoint changes tangential motion only after projection;
$-\eta(t)t g_t$ need not itself be tangent.  The second
term replaces the candidate's normal component with the prescribed residual
attraction.  For $\delta>0$, the multiplier still retains the tangential
component but relaxes the exact normal replacement.  The constrained-velocity
interpretation and regularization effect are derived in
Appendix~\ref{app:multiplier-correction}.

\subsection{Properties and sampling algorithm}
\label{sec:properties-algorithm}

We first identify the exact continuous residual dynamics and the adjoint's
instantaneous effect on constraint-neutral motion, then give the finite-step
sampler and its residual bound.

\begin{algorithm}[!t]
\caption{Unified persistent-anchor MAFM sampling. }
\label{alg:persistent-mafm}
\small
\begin{algorithmic}[1]
\Require source $x_0$, frozen $D_\theta$, constraint $\C$, steps $N$,
schedules $\gamma,\eta,\kappa$, ridge $\delta$
\State \costrev{$h\gets1/N$; \ $x^{(0)}\gets x_0$; \ $d^{(0)}\gets D_\theta(x_0,0)$; \ $a^{(0)}\gets d^{(0)}$}
\For{$k=0,\ldots,N-1$}
  \State \costrev{$t_k\gets kh$; \ $J^{(k)}\gets J_\C(a^{(k)})$}
  \If{\costrev{$k>0$}}
    \State \costrev{$d^{(k)}\gets D_\theta(x^{(k)},t_k)$}
  \EndIf
  \State $q^{(k)}\gets0$
  \If{$\eta(t_k)t_k>0$}
    \State $q^{(k)}\gets\operatorname{clip}_k
    \bigl(\nabla_x\mathcal E_{t_k}(x^{(k)})\bigr)$
    \Comment{one input VJP}
  \EndIf
  \State $w^{(k)}\gets(\rho_k/h)(d^{(k)}-a^{(k)})
    -\eta(t_k)t_kq^{(k)}$ \Comment{tracking and guidance}
  \State solve $\bigl(J^{(k)}(J^{(k)})^\top+\delta I_m\bigr)\lambda^{(k)}
  =-\bigl[J^{(k)}w^{(k)}+\kappa(t_k)\C(a^{(k)})\bigr]$
  \Comment{normal attraction}
  \State $a^{(k+1)}\gets a^{(k)}+h\bigl[w^{(k)}+(J^{(k)})^\top\lambda^{(k)}\bigr]$
  \Comment{advance persistent anchor}
  \State $x^{(k+1)}\gets(1-t_{k+1})x_0+t_{k+1}a^{(k+1)}$
  \Comment{rebuild model query}
\EndFor
\State \Return $x^{(N)}=a^{(N)}$
\end{algorithmic}
\end{algorithm}

\begin{proposition}[Exact residual contraction]
\label{prop:contraction}
Suppose $\C\in C^1$, $J_{\C}(a_t)$ has full row rank along an existing
solution, $\kappa$ is integrable, and Eq.~\eqref{eq:unified-multiplier} is
solved exactly with $\delta=0$.  Then
\begin{equation}
  \frac{\dd}{\dd t}\C(a_t)=-\kappa(t)\C(a_t),\qquad
  \C(a_t)=e^{-\int_0^t\kappa(r)\dd r}\C(a_0).
  \label{eq:contraction}
\end{equation}
\end{proposition}
In this exact setting, the multiplier fixes the residual rate independently of
the candidate velocity: $\kappa$ controls normal attraction, while $\gamma$
and $\eta$ control tangent motion.
Appendix~\ref{app:contraction-proof} proves this result and derives the
additional algebraic terms caused by a ridge or an inexact multiplier solve.

\begin{proposition}[Instantaneous projected-adjoint effect]
\label{prop:adjoint-effect}
Fix a source $x_0$, anchor $a$, and time $t$, and evaluate $g_t$ at the
common query $x=(1-t)x_0+ta$.  Suppose $J_{\C}(a)$ has full row rank and
the multiplier is solved exactly with $\delta=0$.  Let
$V_\eta(a,t;x_0)$ be the corrected anchor velocity from
Eqs.~\eqref{eq:guided-candidate}--\eqref{eq:unified-anchor-ode},
and let $V_0(a,t;x_0)$ use the same
state and schedules with $\eta(t)=0$.  Then
\begin{equation}
  V_\eta-V_0=-\eta(t)tP_T(a)g_t,
  \qquad J_{\C}(a)(V_\eta-V_0)=0.
  \label{eq:projected-adjoint-effect}
\end{equation}
For $t>0$ and $\eta(t)>0$, the adjoint has no instantaneous effect if
and only if $P_T(a)g_t=0$, equivalently
$g_t\in\operatorname{range}(J_{\C}(a)^\top)$.
\end{proposition}
Together, Propositions~\ref{prop:contraction} and~\ref{prop:adjoint-effect}
separate the two roles of the correction: the multiplier prescribes residual
contraction, while the input adjoint can steer constraint-neutral motion through
its tangent projection.  For sparse observations, the endpoint model's learned
cross-coordinate sensitivity can carry information from measured to unmeasured
coordinates through this correction.  Appendix~\ref{app:adjoint-effect} gives
the sparse-observation specialization, and the Poisson ablation in
Section~\ref{sec:core-ablation} examines the adjoint's effect.

Under local common-query-region and Lipschitz assumptions, endpoint and
input-adjoint errors yield a pathwise anchor perturbation bound, stated and
proved in Appendix~\ref{app:endpoint-stability}.

\paragraph{Discrete sampler.}
Algorithm~\ref{alg:persistent-mafm} discretizes the anchor dynamics with
$h=1/N$, $t_k=kh$, and relaxed tracking weight
$\rho_k=1-e^{-\gamma(t_k)h}$.  Its optional $\operatorname{clip}_k$ is the
identity by default; Appendix~\ref{app:trust-clipping} gives the clipped
variant.  The query is rebuilt from the updated anchor as in
Eq.~\eqref{eq:query-state}.
\costrev{Appendix~\ref{app:compute-cost} gives the computational complexity and measured sampling times.}

\begin{proposition}[Finite-step residual control]
\label{prop:discrete-residual}
Consider Algorithm~\ref{alg:persistent-mafm} with $h>0$, $\delta=0$,
full-row-rank $J^{(k)}$, and an exact multiplier solve.  Set
$u^{(k)}=w^{(k)}+(J^{(k)})^\top\lambda^{(k)}$,
$c^{(k)}=\C(a^{(k)})$, and $\kappa_k=\kappa(t_k)$.
Suppose $\C\in C^1$ and $J_{\C}$ is $L_J$-Lipschitz in operator norm
along the step segment.  Then
\begin{equation}
  c^{(k+1)}=(1-h\kappa_k)c^{(k)}+R^{(k)},
  \qquad \|R^{(k)}\|\leq\tfrac12L_Jh^2\|u^{(k)}\|^2.
  \label{eq:discrete-residual}
\end{equation}
\end{proposition}
The proof and the accumulated bound, including regularization and inexact
linear solves, are in Appendix~\ref{app:discrete-residual}.  Under constant
$\kappa>0$, $0<h\kappa\leq1$, bounded step velocities, and exact
unregularized solves, the curvature contribution to the residual bound is
$O(h)$.

%% file: sections/experiments.tex
\input{tables/primary_main}
\section{Experiments}
\label{sec:experiments}

We evaluate \method{} on six physical-space generation problems spanning geometric
constraints, partial observations, conservation laws, and prescribed physical
quantities. The experiments test whether hard constraints can be enforced at
inference time while preserving the physical structure learned by the pretrained
flow model. We first introduce the common experimental setup and evaluation
protocol, and then present benchmark results and additional analyses.

\subsection{Experimental setting and evaluation}
\label{sec:setup}

\paragraph{Benchmarks.}
We evaluate \method{} on six physical-space generation tasks: sphere support
recovery, partially observed Poisson fields, mass-constrained Burgers and
reaction--diffusion trajectories, energy-conditioned Euler flows, and controlled
Navier--Stokes (NS) flows. All methods generate the complete physical state.
Table~\ref{tab:case-overview} in Appendix~\ref{app:task-constraints} distinguishes
the physical structure learned from data from the constraints supplied at inference.

\input{tables/energy_main}

\paragraph{Baselines and metrics.}
We compare vanilla FM with and without terminal projection,
\rev{PCFM}~\citep{utkarsh2026physics}, ECI~\citep{cheng2025gradient},
\rev{stop-gradient guidance~\citep{huang2024diffusionpde}},
\rev{full-gradient endpoint guidance~\citep{yao2026guideddiffusionsamplingfunction}},
\clarrev{CCFM~\citep{liang2025chanceconstrainedflowmatchinghighfidelity}},
\rev{D-Flow~\citep{benhamu2024dflowdifferentiatingflowscontrolled}, PnP-Flow~\citep{martin2025pnpflowplugandplayimagerestoration}, and LDF~\citep{kurtz2026constrainedflowmatchinglagrangian}}.
We evaluate two primary
objectives: constraint satisfaction and fidelity to the learned physical
structure. Constraint satisfaction is measured by the corresponding constraint
error, while physical fidelity is evaluated using PDE relative $L_2$ error.
\rev{For time-dependent PDEs, IC RelL2 compares each sample with an independent rollout initialized from its generated initial field.}
\rev{If the training distribution has a density in the ambient space, it
assigns zero probability to the feasible manifold defined by a regular equality
constraint. Distributional metrics are reported as ancillary diagnostics. Field SWD results appear in Appendix~\ref{app:field-swd}.}
Appendix~\ref{app:protocols} gives feasibility thresholds
and reference construction. Appendix~\ref{app:complete-results} provides sample
counts and full results.

\paragraph{Implementation.}
Within each benchmark, samplers share the frozen flow network, paired starting
sources, supplied conditions, and evaluation set. Experiments were conducted on
a workstation equipped with eight NVIDIA GeForce RTX~4090 GPUs. Appendix~\ref{app:protocols}
details the implementation and hyperparameters. \rev{Appendix~\ref{app:additional-inverse-baselines} describes D-Flow, PnP-Flow, and LDF.}
Appendix~\ref{app:compute-cost} reports inference costs.

\subsection{Geometric and partially observed constraints}
\label{sec:linear}
\paragraph{Learned support under a linear constraint.}
\rev{On sphere--plane, $z=0$ defines the plane and the learned unit-sphere structure selects its circular intersection. Terminal projection satisfies $z=0$ but achieves only 6.47\% joint coverage. MAFM attains mean $|z|=3.37\times10^{-6}$, radial $W_1=0.00493$, and the highest joint coverage (73.0\%). Appendix~\ref{app:complete-results} reports full diagnostics and threshold sensitivity.}

\paragraph{Joint fields under partial observation.}
The Poisson condition observes sparse values of $U$, but the model must
generate a coupled pair $(U,F)$. Figure~\ref{fig:joint-main}(a) shows why
sensor agreement is insufficient: PCFM matches the observations yet leaves
$U$--$F$ inconsistency, while full-gradient guidance reduces the PDE
discrepancy but misses the observation condition. Across $1{,}024$ samples
with 64 sensors, MAFM is feasible and lowers the best feasible PDE RelL2 from
5.55\% to \rev{1.70\%} (Table~\ref{tab:dynamics-main}a). Terminal projection reaches
zero observation error but has 22.7\% PDE RelL2. \clarrev{This pattern is
consistent with input guidance carrying observation information to unobserved
coordinates while normal correction enforces the sensor constraints.} With 256
sensors, MAFM gives the
lowest feasible PDE error (2.25\%; Appendix~\ref{app:complete-results}).

\subsection{Conservation with dynamical consistency}
\label{sec:conservation}
Burgers fixes the spatial mean throughout 65 frames, and reaction--diffusion fixes total two-species mass. \rev{Neither condition determines
how the field evolves. We evaluate physical consistency by computing the
relative $L_2$ error against an independent PDE rollout initialized from each
generated trajectory's initial field. Table~\ref{tab:dynamics-main}b shows that terminal projection meets the mass condition yet yields own-IC errors of 27.0\% and 10.4\%. MAFM attains numerical feasibility with own-IC errors of 0.965\% and 2.74\%, respectively, compared with the next feasible errors of 1.03\% and 2.96\%.}

The ECI$^\ast$ row is operator-equivalent to PCFM-Base on Poisson; it uses
the fixed ECI-2 configuration on Burgers and the registered ECI configuration
on reaction--diffusion. All evaluated variants appear in the appendix.

\subsection{Energy-conditioned prior recovery}
\label{sec:energy}
\begin{revision}
\textcolor{black}{The Euler and controlled-NS samplers receive only a kinetic-energy target;
divergence, spatial mean, and time evolution are not supplied as conditions.
We assess them separately to ask whether the samples still resemble the
physical flows learned from data.} \rev{For each case, we use 512 paired sources and set the
framewise kinetic-energy target to the training energy.}
\textcolor{black}{In the full comparison (Table~\ref{tab:dynamics-main}c),
MAFM has the lowest own-IC, divergence, and mean errors among distinct
energy-feasible methods on both systems. Own-IC error falls from 3.62\% to
2.94\% on Euler and from 3.40\% to 2.76\% on controlled NS relative to the
strongest feasible alternatives.} \textcolor{black}{Figure~\ref{fig:joint-main}(b) shows one controlled-NS sample
per method: generated and own-IC vorticity, their absolute difference, and the
energy residual over time. The velocity relative $L_2$ and mean energy
error values shown in the panels summarize the full held-out set.}

\textcolor{black}{The figure exposes two distinct failure modes. PCFM-Base
nearly eliminates energy error, but its sampled flow departs from its own-IC
rollout. Full-gradient guidance produces a closer rollout but misses the
energy band. MAFM keeps the energy residual within that band while
preserving a closer rollout.} In this setting, the one-mix ECI update is
equivalent to PCFM-Base.

\paragraph{Shifted energy targets.}
We test Euler and controlled NS at 0.8 and 1.2 times the training energy.
MAFM exceeds the energy tolerance before projection. With the shared terminal
projection, it meets the tolerance and achieves the lowest relative $L_2$ error
among feasible methods in all four settings (Appendix~\ref{app:shifted-energy}).
\end{revision}

\subsection{Core-mechanism ablations}
\label{sec:core-ablation}
\input{tables/core_ablation}
\begin{revision}
\textcolor{black}{Table~\ref{tab:core-ablation} compares targeted changes to frozen models
with matched sources, conditions, NFE = 100, and the main feasibility criteria;
no variant is retrained or tuned. Dashes mark task--variant pairs that were not
evaluated.} \costrev{Setting $\eta=0$ removes the input adjoint, and
setting $\gamma=0$ removes endpoint tracking. Both ablations retain normal
attraction. All three Poisson variants meet the observation tolerance with the
same CE ($1.72\times10^{-7}$). The PDE relative $L_2$ error rises from 1.70\%
to 29.0\% without endpoint tracking and to 38.0\% without the input adjoint.}
\textcolor{black}{The latter result is consistent with the role of adjoint
guidance in joint $U$--$F$ consistency in Corollary~\ref{cor:sparse-adjoint}.}
\textcolor{black}{Removing support tracking remains feasible yet increases
own-IC error from 0.965\% to 2.70\% on Burgers and from 2.74\% to 10.6\% on
reaction--diffusion.} \textcolor{black}{Resetting the anchor degrades both feasibility and dynamics,
while removing normal correction makes both systems infeasible.}

\paragraph{What the anchor history contributes.}
For the affine mass constraint, the tangential update blends the previous
anchor with the latest endpoint prediction, carrying information from earlier
predictions into later steps. To test the contribution of this history on
Burgers, we replace the anchor's tangential component at every step with that
of the latest prediction and keep the normal update fixed. The paired per-step
mass residuals differ by at most $3.58\times10^{-7}$, while the relative $L_2$
rollout error rises from 0.965\% to 7.68\%. At comparable mass feasibility, this
result shows that retaining tangential history helps preserve the Burgers
trajectory dynamics. Appendices~\ref{app:anchor-memory} and~\ref{app:anchor-control}
give the recurrence and the full experimental protocol.
\end{revision}

%% file: tables/primary_main.tex
\begin{figure}[t]
\centering
\includegraphics[width=.99\linewidth]{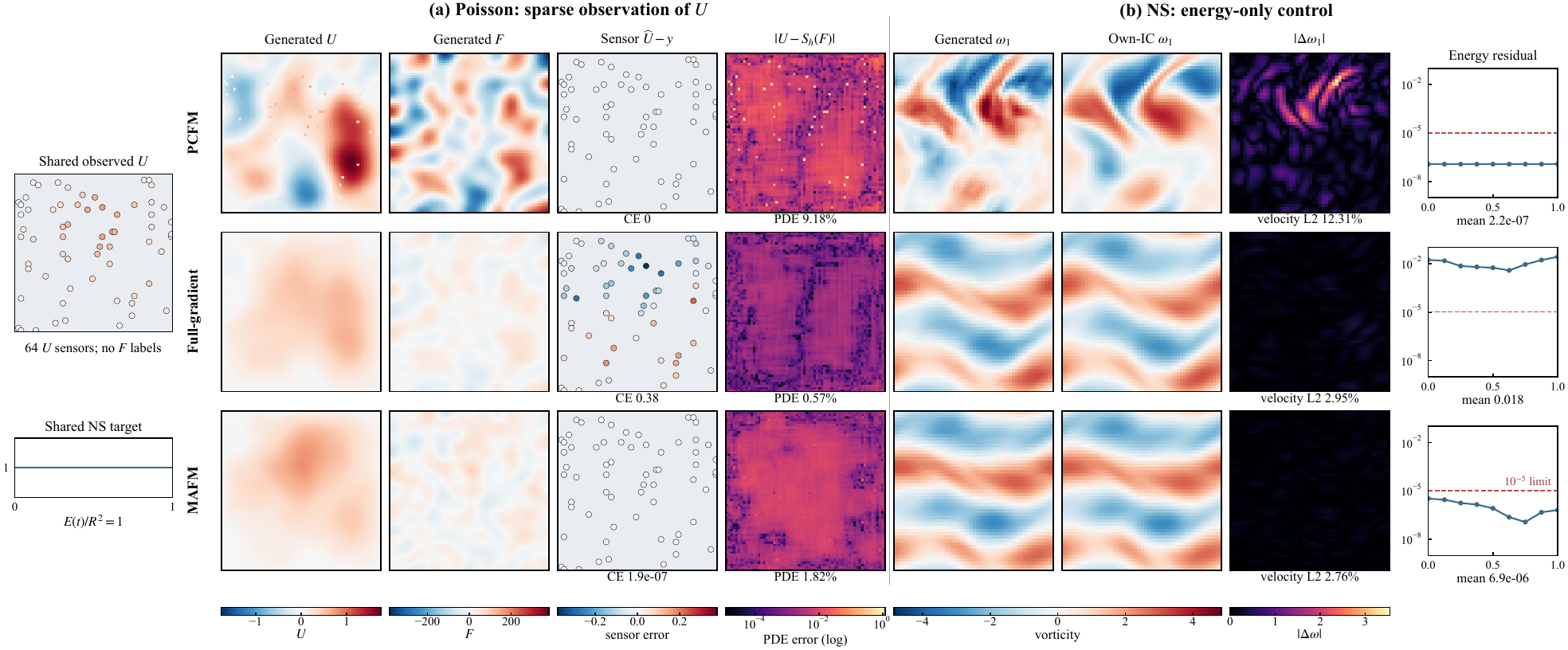}
\caption{\textbf{Representative Poisson and fixed-energy controlled-NS outputs
under shared conditions.}}
\label{fig:joint-main}
\end{figure}

%% file: tables/energy_main.tex
\begin{table}[t]
\renewcommand{\feas}{\textbf{Yes}}
\renewcommand{\infeas}{No}
\caption{\textbf{Physical consistency under supplied constraints.}
(a) Poisson, (b) Burgers and reaction--diffusion, and (c) Euler and controlled NS.}
\label{tab:dynamics-main}
\centering
\scriptsize
\setlength{\tabcolsep}{1.0pt}

\begin{tabular*}{\textwidth}{@{\extracolsep{\fill}}lccccccccc@{}}
\toprule
& \multicolumn{3}{c}{\textbf{(a) Partial observation}}
& \multicolumn{6}{c}{\textbf{(b) Conservation}} \\
\cmidrule(lr){2-4}\cmidrule(l){5-10}
& \multicolumn{3}{c}{\textbf{Poisson, 64 sensors} ($n=1{,}024$)}
& \multicolumn{3}{c}{\textbf{Burgers} ($n=768$)}
& \multicolumn{3}{c}{\textbf{Reaction--diffusion} ($n=384$)} \\
\cmidrule(lr){2-4}\cmidrule(lr){5-7}\cmidrule(l){8-10}
Method & Feas. & CE$\downarrow$ & PDE RelL2$\downarrow$
       & Feas. & CE$\downarrow$ & IC RelL2$\downarrow$
       & Feas. & CE$\downarrow$ & IC RelL2$\downarrow$ \\
\midrule
Vanilla FM & \infeas & 1.40 & 1.13\%
           & \infeas & $3.57\mathrm{e}{-1}$ & 1.13\%
           & \infeas & $5.50\mathrm{e}{-2}$ & 1.31\% \\
FM + projection & \feas & 0 & 22.7\%
                & \feas & $5.49\mathrm{e}{-8}$ & 27.0\%
                & \feas & $5.84\mathrm{e}{-8}$ & 10.4\% \\
PCFM-Base & \feas & 0 & 11.5\%
          & \feas & $5.39\mathrm{e}{-8}$ & 7.61\%
          & \feas & $5.89\mathrm{e}{-8}$ & 3.03\% \\
PCFM-Relaxed & \feas & 0 & 10.3\%
             & \infeas & $3.17\mathrm{e}{-6}$ & 7.61\%
             & \infeas & $3.72\mathrm{e}{-4}$ & 2.46\% \\
ECI$^\ast$ & \feas & 0 & 11.5\%
            & \feas & $5.40\mathrm{e}{-8}$ & 7.54\%
            & \feas & $5.86\mathrm{e}{-8}$ & \second{2.96\%} \\
Stop-grad. & \feas & $3.29\mathrm{e}{-9}$ & \second{5.55\%}
           & \feas & $5.38\mathrm{e}{-8}$ & \second{1.03\%}
                    & \feas & $5.92\mathrm{e}{-8}$ & 6.04\% \\
Full-grad. (DPDE) & \infeas & 0.610 & 1.05\%
                   & \infeas & $2.63\mathrm{e}{-2}$ & 17.2\%
                    & \infeas & $1.70\mathrm{e}{-3}$ & 1.56\% \\
\clarrev{CCFM ($n=2$)} & \clarrev{\feas} & \clarrev{0} & \clarrev{22.78\%}
  & \clarrev{\feas} & \clarrev{$5.39\mathrm{e}{-8}$} & \clarrev{5.73\%}
  & \clarrev{\feas} & \clarrev{$5.96\mathrm{e}{-8}$} & \clarrev{6.83\%} \\
\rev{D-Flow}
& \rev{\infeas} & \rev{0.555} & \rev{1.03\%}
& \rev{\infeas} & \rev{0.995} & \rev{39.01\%}
& \rev{\infeas} & \rev{0.0317} & \rev{3.45\%} \\
\rev{PnP-Flow}
& \rev{\infeas} & \rev{0.538} & \rev{1.22\%}
& \rev{\infeas} & \rev{$1.49\mathrm{e}{-3}$} & \rev{0.71\%}
& \rev{\infeas} & \rev{$2.21\mathrm{e}{-3}$} & \rev{0.98\%} \\
\rev{LDF (198 NFE)}
& \rev{\infeas} & \rev{$3.84\mathrm{e}{-2}$} & \rev{14.96\%}
& \rev{\infeas} & \rev{$7.22\mathrm{e}{-3}$} & \rev{4.54\%}
& \rev{\infeas} & \rev{$2.69\mathrm{e}{-3}$} & \rev{21.81\%} \\
\textbf{MAFM} & \feas & $1.72\mathrm{e}{-7}$ & \rev{\best{1.70\%}}
               & \feas & $4.22\mathrm{e}{-8}$ & \best{0.965\%}
               & \feas & $1.13\mathrm{e}{-7}$ & \best{2.74\%} \\
\quad + projection & \feas & 0 & \rev{\best{1.70\%}}
                   & \feas & $5.32\mathrm{e}{-8}$ & \best{0.965\%}
                   & \feas & $6.77\mathrm{e}{-8}$ & \best{2.74\%} \\
\bottomrule
\end{tabular*}

\vspace{4pt}

\setlength{\tabcolsep}{1.4pt}
\begin{tabular*}{\textwidth}{@{\extracolsep{\fill}}lcccccccccc@{}}
\toprule
\multicolumn{11}{c}{\textbf{(c) Energy-only held-out tests} ($512$ samples per system)} \\
\midrule
& \multicolumn{5}{c}{\textbf{Euler}}
& \multicolumn{5}{c}{\textbf{Controlled NS}} \\
\cmidrule(lr){2-6}\cmidrule(l){7-11}
Method & Feas. & E$\downarrow$ & IC RelL2$\downarrow$ & Div.$\downarrow$ & Mean$\downarrow$
       & Feas. & E$\downarrow$ & IC RelL2$\downarrow$ & Div.$\downarrow$ & Mean$\downarrow$ \\
\midrule
Vanilla FM
& \infeas & $1.85\mathrm{e}{-2}$ & 3.68\% & 0.0635 & 0.0131
& \infeas & $1.80\mathrm{e}{-2}$ & 3.46\% & 0.0624 & 0.0111 \\
Terminal projection
& \feas & $2.29\mathrm{e}{-7}$ & \second{3.62\%} & \second{0.0636} & 0.0131
& \feas & $2.32\mathrm{e}{-7}$ & \second{3.40\%} & \second{0.0624} & 0.0113 \\
\rev{PCFM}
& \infeas & $1.84\mathrm{e}{-3}$ & 13.6\% & 0.229 & 0.0297
& \infeas & $1.81\mathrm{e}{-3}$ & 12.3\% & 0.212 & 0.0301 \\
\rev{\quad + projection}
& \rev{\feas} & \rev{$2.34\mathrm{e}{-7}$} & \rev{13.6\%} & \rev{0.2287} & \rev{0.0298}
& \rev{\feas} & \rev{$2.44\mathrm{e}{-7}$} & \rev{12.3\%} & \rev{0.2124} & \rev{0.0302} \\
\rev{ECI}
& \feas & $2.12\mathrm{e}{-7}$ & 13.6\% & 0.232 & 0.0276
& \feas & $2.21\mathrm{e}{-7}$ & 12.3\% & 0.213 & 0.0301 \\
Stop-grad. guidance
& \infeas & $2.13\mathrm{e}{-2}$ & 8.64\% & 0.112 & 0.0214
& \infeas & $1.97\mathrm{e}{-2}$ & 3.47\% & 0.0494 & 0.00893 \\
Full-grad. (DPDE)
& \infeas & $1.60\mathrm{e}{-2}$ & 3.42\% & 0.0604 & 0.0106
& \infeas & $1.78\mathrm{e}{-2}$ & 2.95\% & 0.0515 & 0.00729 \\
\clarrev{CCFM ($n=2$)}
& \clarrev{\feas} & \clarrev{$2.33\mathrm{e}{-7}$} & \clarrev{3.68\%} & \clarrev{0.0650} & \clarrev{\second{0.0130}}
& \clarrev{\feas} & \clarrev{$2.35\mathrm{e}{-7}$} & \clarrev{3.48\%} & \clarrev{0.0640} & \clarrev{\second{0.0106}} \\
\rev{D-Flow}
& \rev{\infeas} & \rev{$7.51\mathrm{e}{-3}$} & \rev{3.54\%} & \rev{0.0626} & \rev{0.0108}
& \rev{\infeas} & \rev{$7.48\mathrm{e}{-3}$} & \rev{3.27\%} & \rev{0.0599} & \rev{0.00941} \\
\rev{PnP-Flow}
& \rev{\infeas} & \rev{0.0581} & \rev{11.33\%} & \rev{0.203} & \rev{0.0300}
& \rev{\infeas} & \rev{0.0554} & \rev{10.25\%} & \rev{0.189} & \rev{0.0274} \\
\rev{LDF (198 NFE)}
& \rev{\infeas} & \rev{$5.85\mathrm{e}{-4}$} & \rev{7.13\%} & \rev{0.165} & \rev{0.0219}
& \rev{\infeas} & \rev{$5.35\mathrm{e}{-4}$} & \rev{5.86\%} & \rev{0.152} & \rev{0.0184} \\
\textbf{MAFM}
& \feas & $6.84\mathrm{e}{-6}$ & \best{2.94\%} & \best{0.0483} & \best{0.0108}
& \feas & $6.89\mathrm{e}{-6}$ & \best{2.76\%} & \best{0.0474} & \best{0.00881} \\
\quad + projection
& \feas & $2.37\mathrm{e}{-7}$ & \best{2.94\%} & \best{0.0483} & \best{0.0108}
& \feas & $2.43\mathrm{e}{-7}$ & \best{2.76\%} & \best{0.0474} & \best{0.00881} \\
\bottomrule
\end{tabular*}
\end{table}

%% file: tables/core_ablation.tex
\begin{table}[t]
\renewcommand{\feas}{\textbf{Yes}}
\renewcommand{\infeas}{No}
\caption{\textbf{Core-mechanism ablations.}}
\label{tab:core-ablation}
\centering
\scriptsize
\setlength{\tabcolsep}{3pt}
\renewcommand{\arraystretch}{1.04}

\begin{tabular*}{\textwidth}{@{\extracolsep{\fill}}lccccccc@{}}
\toprule
& \multicolumn{1}{c}{\textbf{(a) Poisson} ($n=1{,}024$)}
& \multicolumn{6}{c}{\textbf{(b) Conservation}} \\
\cmidrule(lr){2-2}\cmidrule(l){3-8}
& & \multicolumn{3}{c}{\textbf{Burgers} ($n=768$)}
& \multicolumn{3}{c}{\textbf{Reaction--diffusion} ($n=384$)} \\
\cmidrule(lr){3-5}\cmidrule(l){6-8}
Variant & PDE RelL2$\downarrow$
        & Feas. & CE$\downarrow$ & IC RelL2$\downarrow$
        & Feas. & CE$\downarrow$ & IC RelL2$\downarrow$ \\
\midrule
Full MAFM & \best{1.70\%}
          & \feas & $4.22\mathrm{e}{-8}$ & \best{0.965\%}
          & \feas & $1.13\mathrm{e}{-7}$ & \best{2.74\%} \\
w/o adjoint & 38.0\%
            & \multicolumn{3}{c}{\textemdash}
            & \multicolumn{3}{c}{\textemdash} \\
Reset anchor & \textemdash
             & \infeas & $9.00\mathrm{e}{-3}$ & 9.44\%
             & \infeas & $4.41\mathrm{e}{-2}$ & 6.14\% \\
w/o support & \costrev{29.0\%}
            & \feas & $1.18\mathrm{e}{-7}$ & 2.70\%
            & \feas & $1.15\mathrm{e}{-7}$ & 10.6\% \\
w/o normal & \textemdash
           & \infeas & 0.270 & 1.29\%
           & \infeas & $3.19\mathrm{e}{-2}$ & 2.19\% \\
\bottomrule
\end{tabular*}
\end{table}

%% file: sections/conclusion.tex
\section{Conclusion}
\label{sec:conclusion}
We study inference-time sampling from the intersection of a
learned data support and an equality constraint supplied only at generation
time. MAFM addresses this problem through the continuous evolution of a
persistent physical anchor driven by a frozen endpoint model. Successive
endpoint predictions are incorporated through finite-rate tracking, while a
geometric correction separates model-guided tangent motion from normal
constraint attraction. Under full-rank Jacobians and exact unregularized
multiplier solves, the resulting dynamics yield a prescribed
residual-contraction law, while the persistent-anchor relaxation admits a
local feedback-stabilization analysis.

Across six benchmarks, MAFM achieves numerical feasibility while maintaining
strong consistency with the physical structure learned by the frozen model.
The ablations further support the complementary roles of persistent tracking,
tangent guidance, and normal correction. While MAFM provides explicit control
over constraint-residual dynamics, specifying and matching a particular
probability law on the feasible intersection remains a separate problem.
Extending the framework toward distributionally controlled constrained
sampling and broader physical regimes is an important direction for future
work.

%% file: sections/statements.tex
\section*{Reproducibility Statement}
The main paper
provides the target definition, assumptions, sampler equations, pseudocode,
and key experimental results, while Appendix~\ref{app:proofs} presents the
theoretical derivations and proofs.
Appendix~\ref{app:protocols} describes the experimental protocols, including
physical tensors, constraints, and evaluation metrics. Appendix~\ref{app:experiments}
provides the complete result tables and projection audits.
We also record code and input hashes, logs, checkpoints, generated samples,
and evaluation outputs, and clearly distinguish the single training seed from
the number of generated samples. Negative or mixed outcomes are reported
alongside improvements. Together, these materials provide the information and
implementation needed for independent verification and reproduction of our
reported results.

\section*{Ethics Statement}
The work uses procedurally generated geometric and physical data and involves
no personal data or human participants.  A relevant downstream risk is treating
a small declared residual as certification that a sample follows the correct
physical distribution or governing dynamics.  We therefore evaluate these
properties separately and recommend independent verification for scientific or
safety-critical use.

\section*{AI Use Statement}
Generative AI tools assisted with language editing, translation, manuscript
organization, LaTeX assembly, implementation of supplementary comparison
scripts, and interpretation of archived results. The authors take
responsibility for the final manuscript and its scientific claims.

\section*{Data, Conflicts, and Funding}
All experiments use synthetic data described in the paper and appendix.  The
authors declare no competing interests and no external funding that
creates a conflict of interest.

%% file: sections/appendix.tex
\section*{Appendix contents}
\noindent\hyperref[app:proofs]{A. Proofs and derivations}\hfill\pageref{app:proofs}\\
\hyperref[app:algorithms]{B. Additional algorithmic details}\hfill\pageref{app:algorithms}\\
\hyperref[app:protocols]{C. Experimental protocols}\hfill\pageref{app:protocols}\\
\hyperref[app:complete-results]{D. Supplementary results and diagnostics}\hfill\pageref{app:complete-results}\\
\hyperref[app:numerical-boundaries]{E. Numerical and applicability boundaries}\hfill\pageref{app:numerical-boundaries}
\par\smallskip
\noindent\textit{Cases in Appendix C:} \hyperref[app:sphere-protocol]{sphere--plane},
\hyperref[app:poisson-protocol]{Poisson},
\hyperref[app:burgers-protocol]{Burgers},
\hyperref[app:rd-protocol]{reaction--diffusion},
\hyperref[app:euler-protocol]{Euler}, and
\hyperref[app:ns-protocol]{controlled NS}.

\section{Proofs and derivations}
\label{app:proofs}

\subsection{Target-measure geometry}
\label{app:target-geometry}

The main text specifies the support of $\mu_q$ through
$\M=\Sdata\cap\Ncon$.  A distributional target requires an additional
choice of weights on that intersection.  For a measurable set $B$, one such
choice is $p_{\mathrm{data}}(B\mid\Ncon)$ when
$p_{\mathrm{data}}(\Ncon)>0$.  For a zero-probability constraint event, fix
the residual norm and normalization and, when the denominator is positive and
the weak limit exists, define the tubular conditional law
\begin{equation}
  \nu_\varepsilon(B)=
  \frac{p_{\mathrm{data}}\!\left(B\cap
    \{x:\|\C(x)\|\leq\varepsilon\}\right)}
       {p_{\mathrm{data}}\!\left(
    \{x:\|\C(x)\|\leq\varepsilon\}\right)},
  \qquad \nu=\lim_{\varepsilon\downarrow0}\nu_\varepsilon.
  \label{eq:tube-law}
\end{equation}
Suppose $\Sdata$ is a smooth $d_{\Sdata}$-dimensional manifold with
$d_{\Sdata}\geq m$, $\C|_{\Sdata}$ has full row rank near $\M$, and
$p_{\mathrm{data}}$ has density $p_{\Sdata}$ with respect to
$\mathcal H^{d_{\Sdata}}$.  Subject to the regularity and integrability
conditions for the normalized limit, the coarea formula gives
\begin{equation}
  \dd\nu(x)\propto
  \frac{p_{\Sdata}(x)}
  {\sqrt{\det\!\left(J_{\Sdata}(x)J_{\Sdata}(x)^\top\right)}}
  \,\dd\mathcal H^{d_{\Sdata}-m}(x),
  \qquad x\in\M,
  \label{eq:coarea-target}
\end{equation}
where $J_{\Sdata}$ is the Jacobian of $\C$ in an orthonormal tangent basis
of $T_x\Sdata$.  The tube law depends on the chosen residual units and norm;
reparameterizing the same zero set may change its limiting weights.

Let $\nu_{\rm MAFM}^*=\operatorname{Law}(a_1^*)$ be the terminal law of the
ideal anchor dynamics and $\widehat a^{(N)}$ a numerical output.  When these
laws and $\nu$ have finite second moments, the Wasserstein triangle
inequality gives
\begin{equation}
\begin{aligned}
W_2(\operatorname{Law}(\widehat a^{(N)}),\nu)
\leq{}&W_2(\operatorname{Law}(\widehat a^{(N)}),\nu_{\rm MAFM}^*)\\
&+W_2(\nu_{\rm MAFM}^*,\nu).
\end{aligned}
\label{eq:target-law-bound}
\end{equation}
The first term can reflect model and numerical errors; separating it further
requires assumptions linking their coupled effects.  The second term is the
structural difference between ideal anchor dynamics and the additionally
specified target law.

\subsection{Endpoint estimation and query-region conditions}
\label{app:endpoint-details}

\paragraph{Conditional mean and regression error.}

For the linear bridge in Eq.~\eqref{eq:linear-path}, the population flow field
is $v^*(x,t)=\mathbb E[x_1-x_0\mid x_t^{\mathrm{train}}=x]$.  The
pointwise identity
$x_t^{\mathrm{train}}+(1-t)(x_1-x_0)=x_1$ implies, after
conditioning on $x_t^{\mathrm{train}}=x$,
\begin{equation}
  x+(1-t)v^*(x,t)=\mathbb E[x_1\mid x_t^{\mathrm{train}}=x].
\end{equation}
The conditional expectation is also the $L^2$-optimal endpoint predictor.  For
finite second moments, its orthogonality decomposition is
\begin{equation}
\begin{aligned}
\mathbb E_\pi\|D_\theta(x_t^{\mathrm{train}},t)-x_1\|^2
={}&\mathbb E_\pi\|D_\theta(x_t^{\mathrm{train}},t)-D^*(x_t^{\mathrm{train}},t)\|^2\\
&+\mathbb E_\pi\|x_1-D^*(x_t^{\mathrm{train}},t)\|^2.
\end{aligned}
\label{eq:endpoint-error-decomp}
\end{equation}
The second term quantifies uncertainty about a paired endpoint even for an
optimal predictor.  Under an independent coupling at $t=0$,
$D^*(x_0,0)=\mathbb E[x_1]$, which can fall between data modes.

\paragraph{Endpoint extrapolation and ODE integration.}
Let $\chi_r$ solve the frozen Vanilla ODE from $\chi_t=x$.  Its terminal
state and the endpoint extrapolation are
\begin{equation}
\chi_1=x+\int_t^1v_\theta(\chi_r,r)\dd r,
\qquad D_\theta(x,t)=x+(1-t)v_\theta(x,t).
\label{eq:ode-endpoint-comparison}
\end{equation}
If the velocity along this trajectory is differentiable with
$\|\dd v_\theta(\chi_r,r)/\dd r\|\leq A$ for $r\in[t,1]$, then
\begin{equation}
\|\chi_1-D_\theta(x,t)\|\leq\tfrac12A(1-t)^2.
\label{eq:ode-endpoint-error}
\end{equation}
This local extrapolation bound depends on the stated trajectory regularity.
PCFM also uses endpoint extrapolation \citep{utkarsh2026physics}; the persistent
MAFM anchor incorporates successive predictions across changing queries.

\paragraph{Parameterization and source law.}
For $t<1$, a directly trained endpoint predictor induces the velocity
$v_\theta(x,t)=[D_\theta(x,t)-x]/(1-t)$.  The two parameterizations have
corresponding population optima under the linear path, while their
cross-time finite-model losses carry different $(1-t)^2$ weights.  The
conditional-mean identity requires finite moments and the linear path, rather
than a standard-Gaussian source.  More general flow matching permits other
source laws \citep{tong2023improving}.  MAFM draws $x_0$ from the source
distribution used to train the frozen Vanilla model; task-specific Gaussian
covariances are recorded with their training and sampling protocols.

\paragraph{Query region.}
For a fixed $x_0$, let $\mathcal A_t$ contain the compared anchor paths and
define $\Omega_t=\{(1-t)x_0+ta:a\in\mathcal A_t\}$.  The stability result
uses endpoint and adjoint error bounds on this common query region.  These
are local model assumptions: endpoint regression on the training path alone
does not supply uniform error or Jacobian bounds over every later query.

\subsection{Lagrange-multiplier correction}
\label{app:multiplier-correction}

At a fixed anchor and time, write $J=J_{\C}(a)$ and $c=\C(a)$, and let
$w=w_t$ be the candidate in Eq.~\eqref{eq:guided-candidate}.  The base
specialization sets $\eta=0$.  Under full row rank,
the exact unregularized velocity is the unique solution of
\begin{equation}
  \min_{u\in\mathbb R^d}\tfrac12\|u-w\|^2
  \quad\text{subject to}\quad Ju=-\kappa c.
  \label{eq:velocity-projection}
\end{equation}
For the Lagrangian
$\mathcal L(u,\lambda)=\tfrac12\|u-w\|^2-\lambda^\top(Ju+\kappa c)$,
stationarity gives $u=w+J^\top\lambda$.  Substituting into the constraint
gives $JJ^\top\lambda=-(Jw+\kappa c)$, which yields the main-text
multiplier system in Eq.~\eqref{eq:unified-multiplier}.  The correction
lies in $\operatorname{range}(J^\top)$ and retains the tangent component of
the candidate velocity.

For $\delta>0$, the regularized system instead minimizes
\begin{equation}
  \tfrac12\|u-w\|^2+\frac{1}{2\delta}\|Ju+\kappa c\|^2.
  \label{eq:regularized-velocity-projection}
\end{equation}
Stationarity is $u-w+\delta^{-1}J^\top(Ju+\kappa c)=0$.
Setting $\lambda=-\delta^{-1}(Ju+\kappa c)$ gives
$(JJ^\top+\delta I_m)\lambda=-(Jw+\kappa c)$ and
$u=w+J^\top\lambda$.  Regularization therefore implements a quadratic
penalty on the prescribed residual rate.

\paragraph{Base-specialization endpoint filtering.}
Let $V_D(a,t;x_0)$ denote the exact unregularized base-MAFM velocity using
endpoint field $D$.  At the same anchor, source, and time,
\begin{equation}
  V_{D+e}(a,t;x_0)-V_D(a,t;x_0)
  =\gamma(t)P_T(a)e(x,t),\qquad x=(1-t)x_0+ta.
  \label{eq:projected-endpoint-error}
\end{equation}
For the pointwise perturbation in Eq.~\eqref{eq:projected-endpoint-error},
the two velocities share $a$, $t$, $x_0$, and hence $J$, $P_T$, and $\C(a)$.
Subtracting their base-specialization forms in
Eq.~\eqref{eq:unified-tangent-normal} leaves
$\gamma(t)P_T(a)e(x,t)$.  In particular, errors in
$\operatorname{range}(J_{\C}(a)^\top)$ have no instantaneous effect on the velocity.
With regularization $\delta>0$, $P_T$ is replaced by
$A_\delta=I_d-J^\top(G+\delta I_m)^{-1}J$: normal errors are attenuated rather
than eliminated.  Its tangent eigenvalues are one and its normal eigenvalues
are $\delta/(\sigma_i(J)^2+\delta)$, so $\|A_\delta\|\leq1$.

\subsection{Projected-adjoint effect and sparse observations}
\label{app:adjoint-effect}

\begin{proof}[Proof of Proposition~\ref{prop:adjoint-effect}]
At the fixed anchor and time in the proposition, write
$J=J_{\C}(a)$, $G=JJ^\top$, and
$P_T=I_d-J^\top G^{-1}J$.  The candidate velocities with and without
the adjoint satisfy
\begin{equation}
  w_t^{(\eta)}=\gamma(t)[D_\theta(x,t)-a]-\eta(t)t g_t,
  \qquad w_t^{(0)}=\gamma(t)[D_\theta(x,t)-a].
\end{equation}
Equation~\eqref{eq:unified-tangent-normal} gives
$V_\eta=P_Tw_t^{(\eta)}-J^\top G^{-1}\kappa(t)\C(a)$ and the same
expression with $w_t^{(0)}$ for $V_0$.  Subtracting cancels the normal
attraction term and yields
$V_\eta-V_0=-\eta(t)tP_Tg_t$.  Since $JP_T=0$,
this difference induces no first-order residual change.  Moreover,
$P_T$ is the orthogonal projector onto $\ker J$, with
$\ker P_T=\operatorname{range}(J^\top)$.  For $t>0$ and $\eta(t)>0$,
the velocity difference therefore vanishes exactly when
$g_t\in\operatorname{range}(J^\top)$.
\end{proof}

\begin{corollary}[Sparse-observation pullback]
\label{cor:sparse-adjoint}
After reordering coordinates, let $\C(z)=Hz-y$ with $H=[I_m\;0]$,
partition the physical query
as $x=(x_O,x_U)$.  Write
$D_O(x,t)=HD_\theta(x,t)$.  Then
\begin{equation}
  P_Tg_t=
  \begin{bmatrix}
    0\\
    J_{D_O,x_U}(x_t,t)^\top
    \bigl[D_O(x_t,t)-y\bigr]
  \end{bmatrix}.
  \label{eq:sparse-adjoint-transfer}
\end{equation}
Thus the adjoint can affect unobserved coordinates through the sensitivity
of the predicted observed endpoint to those coordinates.
\end{corollary}
\begin{proof}
Here $HH^\top=I_m$ and
$P_T=I_d-H^\top H=\operatorname{diag}(0,I_{d-m})$.  The input adjoint is
$g_t=J_{D_O}(x_t,t)^\top[D_O(x_t,t)-y]$.
Partitioning this gradient into its $x_O$ and $x_U$ blocks and applying
$P_T$ gives Eq.~\eqref{eq:sparse-adjoint-transfer}.
\end{proof}
For the Poisson state $x=(U,F)$, $x_U$ in this corollary contains both
unobserved $U$ coordinates and the forcing field $F$.  A nonzero
$J_{D_O,x_F}^\top[D_O-y]$ provides a model-mediated route from an
observed-$U$ residual to a tangent update of $F$.  The ablation in
Section~\ref{sec:core-ablation} is consistent with this mechanism; the
identity alone does not guarantee improved joint consistency.

\paragraph{Scope and diagnostics.}
At a common anchor, let $J=J_{\C}(a)$, $G=JJ^\top$, and
$A_\delta=I_d-J^\top(G+\delta I_m)^{-1}J$.  For an exact regularized
multiplier solve with $\delta>0$, the same fixed-state comparison gives
\begin{equation}
  V_{\eta,\delta}-V_{0,\delta}
  =-\eta(t)tA_\delta g_t,
  \qquad
  J(V_{\eta,\delta}-V_{0,\delta})
  =-\eta(t)t\delta(G+\delta I_m)^{-1}Jg_t.
  \label{eq:regularized-adjoint-effect}
\end{equation}
Thus regularization does not fully remove the adjoint's normal component.
For a finite step with nonlinear $\C$, curvature can also make the
next-step residuals differ, as in Eq.~\eqref{eq:discrete-residual}.

The per-unit-$\eta$ instantaneous speed is
$A_t=t\|P_T(a_t)g_t\|$.  A relative diagnostic, when the base
tracking speed is appreciable, is
\begin{equation}
  R_t=\frac{\eta(t)A_t}
  {\gamma(t)\|P_T(a_t)[D_\theta(x_t,t)-a_t]\|+\varepsilon},
  \qquad \varepsilon>0.
  \label{eq:adjoint-activity-score}
\end{equation}
The denominator uses base tracking rather than the guided $P_Tw_t$.
Neither score establishes that guidance is beneficial.  In the discrete
sampler, the realized tangent displacement also includes $h$; this
quantity should be logged when assessing adjoint activity.

\subsection{History weights for affine constraints}
\label{app:anchor-memory}

For $\C(a)=Ha-y$ with full-row-rank $H$, the constant tangent projector
$P_T=I_d-H^\top(HH^\top)^{-1}H$ satisfies $P_TH^\top=0$.
Applying it to Algorithm~\ref{alg:persistent-mafm} removes the multiplier
correction.  For the base specialization $\eta=0$, an exact unregularized
update gives
\begin{equation}
  \begin{aligned}
    P_Ta^{(k+1)}&=(1-\rho_k)P_Ta^{(k)}+\rho_kP_Td^{(k)},\\
    c^{(k+1)}&=(1-h\kappa_k)c^{(k)},
  \end{aligned}
  \qquad c^{(k)}=\C(a^{(k)}),\quad \kappa_k=\kappa(t_k).
  \label{eq:anchor-memory}
\end{equation}
The tangential identity holds for any multiplier values, including
regularized or inexact solves.  For the normal identity, the exact
unregularized solve gives
$H(w^{(k)}+H^\top\lambda^{(k)})=-\kappa_kc^{(k)}$;
affinity then gives the stated recurrence.  Writing
$u^{(k)}=P_Ta^{(k)}$ and $d_T^{(k)}=P_Td^{(k)}$, repeated substitution gives
\begin{equation}
  u^{(N)}=\left[\prod_{j=0}^{N-1}(1-\rho_j)\right]u^{(0)}
  +\sum_{i=0}^{N-1}\left[\rho_i\prod_{j=i+1}^{N-1}(1-\rho_j)\right]d_T^{(i)}.
  \label{eq:anchor-history-weights}
\end{equation}
Empty products equal one.  For $\gamma(t_k)\geq0$, all weights are
nonnegative and sum to one.  The initial-state weight is
$\exp[-h\sum_{j=0}^{N-1}\gamma(t_j)]$.  This makes the role of the attraction
schedule explicit: it determines how much past information remains in the
tangential anchor.  The endpoint estimates themselves depend on evolving
queries.  Active adjoint guidance adds
$-h\eta(t_k)t_kP_Tq^{(k)}$ to the tangent recurrence, with $q^{(k)}$
as in Algorithm~\ref{alg:persistent-mafm}; the normal recurrence remains the
one in Eq.~\eqref{eq:anchor-memory}.

\begin{revision}
\paragraph{Relation to relaxed endpoint projection.}
For an affine equality, let $\hat d^{(k)}=\Pi_{\Ncon}(d^{(k)})$ denote the
orthogonal projection of the current endpoint prediction. A damped
projection moves the prediction partway toward $\hat d^{(k)}$,
giving $z=(1-\alpha)d^{(k)}+\alpha\hat d^{(k)}$ with $0\leq\alpha\leq1$.
Its residual is $\C(z)=(1-\alpha)\C(d^{(k)})$, and its tangential component
is $P_Tz=P_Td^{(k)}$. The damping controls how much of the current
prediction's residual is removed.

A relaxed iteration on the anchor blends the previous anchor with the
projected prediction, $a^{(k+1)}=(1-\alpha)a^{(k)}+\alpha\hat d^{(k)}$.
Since $\C(\hat d^{(k)})=0$ and $P_T\hat d^{(k)}=P_Td^{(k)}$, this gives
\[
  P_Ta^{(k+1)}=(1-\alpha)P_Ta^{(k)}+\alpha P_Td^{(k)},
  \qquad c^{(k+1)}=(1-\alpha)c^{(k)}.
\]
Here the same weight controls how much new tangential information enters
the anchor and how much of its residual is removed. In the exact,
unregularized MAFM step with $\eta=0$ in Eq.~\eqref{eq:anchor-memory},
these fractions are $\rho_k$ and $h\kappa_k$ and can be set independently.
For instance, $0<\rho_k<1$ and $h\kappa_k=1$ retain part of the previous
tangential anchor while removing the full affine residual in that step.
\end{revision}

\rev{The affine recurrence also makes the PCFM connection in the main text
explicit. If $\eta=0$, $\rho_k=1$, and $h\kappa_k=1$ with an exact,
unregularized solve, the next anchor has tangential component $P_Td^{(k)}$
and satisfies $Ha^{(k+1)}=y$. Thus
$a^{(k+1)}=P_Td^{(k)}+H^\dagger y=d^{(k)}-H^\dagger(Hd^{(k)}-y)$,
where $H^\dagger=H^\top(HH^\top)^{-1}$. This is the orthogonal endpoint
projection for an affine equality. Interpolation with the same fixed source
then gives the corresponding PCFM query update.}

\subsection{Proof of Proposition~\ref{prop:contraction}}
\label{app:contraction-proof}

By the chain rule and Eq.~\eqref{eq:unified-anchor-ode},
\begin{align}
  \frac{\dd}{\dd t}\C(a_t)
  &=J_{\C}(a_t)\dot a_t\\
  &=J_{\C}(a_t)w_t+G(a_t)\lambda_t.
\end{align}
With $\delta=0$ and the exact multiplier solution,
$G(a_t)\lambda_t=-[J_{\C}(a_t)w_t+\kappa(t)\C(a_t)]$.  Substitution yields
$\frac{\dd}{\dd t}\C(a_t)=-\kappa(t)\C(a_t)$.  Solving this linear
vector ODE gives Eq.~\eqref{eq:contraction}.

For $\delta>0$, use
$\lambda=-(G+\delta I_m)^{-1}(J_{\C}w+\kappa\C)$ and
$I_m-G(G+\delta I_m)^{-1}=\delta(G+\delta I_m)^{-1}$ to obtain
\begin{equation}
  \frac{\dd}{\dd t}\C(a_t)
  =\delta(G+\delta I_m)^{-1}J_{\C}w_t
  -\kappa(t)G(G+\delta I_m)^{-1}\C(a_t).
  \label{eq:regularized-residual}
\end{equation}
If the numerical multiplier leaves
algebraic residual
$r_\lambda=(G+\delta I_m)\widehat\lambda+J_{\C}w+\kappa\C$,
then $G(G+\delta I_m)^{-1}r_\lambda$ adds a further forcing term to the residual
dynamics.

\subsection{Base-specialization local feedback stability}
\label{app:feedback-stability}

To isolate anchor memory, fix a source $x_0$, time $\bar t$, and affine
constraint $Ha=y$.  Write $a=p+a_\perp$, with $p\in T=\ker H$ and
$a_\perp=H^\top(HH^\top)^{-1}y$, and define
$F(p)=P_TD_\theta((1-\bar t)x_0+\bar t(p+a_\perp),\bar t)$.
Direct replacement uses $p^+=F(p)$, while the relaxed persistent anchor uses
$p^+=(1-\rho)p+\rho F(p)$.

\begin{proposition}[Local feedback stabilization]
\label{prop:feedback-stability}
Suppose $F:T\to T$ is continuously differentiable near a fixed point $p^*$,
and let $K=\mathrm DF(p^*)$.  For $0<\rho\leq1$, the relaxed update has
linearization $M_\rho=(1-\rho)I_T+\rho K$, with eigenvalues
\begin{equation}
  \Lambda_\rho(\zeta)=1-\rho+\rho\zeta,
  \qquad \zeta\in\operatorname{spec}(K).
  \label{eq:feedback-spectrum}
\end{equation}
The fixed point is locally asymptotically stable if
$\max_{\zeta\in\operatorname{spec}(K)}|\Lambda_\rho(\zeta)|<1$.
In particular, a real mode $\zeta<-1$ is damped when
$0<\rho<2/(1-\zeta)$.
\end{proposition}

\begin{proof}
Fix $\bar t$, $x_0$, and the feasible affine slice as in the proposition.
For $a=p+a_\perp$, orthogonal projection of an endpoint onto $Ha=y$ is
$a_\perp+P_TD_\theta((1-\bar t)x_0+\bar t a,\bar t)$.
Thus its tangential direct-replacement map is $F$, and the frozen-time
MAFM map is $H_\rho(p)=(1-\rho)p+\rho F(p)$.
For $\rho>0$, $H_\rho(p)=p$ if and only if $F(p)=p$.
Differentiating at $p^*$ gives
$\mathrm DH_\rho(p^*)=(1-\rho)I_T+\rho K$.
When the endpoint estimator is differentiable at the associated query $x^*$,
$K=\bar t\,P_TJ_{D_\theta}(x^*,\bar t)|_T$.
The spectral mapping identity for this affine polynomial in $K$ gives
Eq.~\eqref{eq:feedback-spectrum}.

If every eigenvalue of $M_\rho$ has modulus less than one, there is an
equivalent norm on $T$ whose induced matrix norm satisfies
$\|M_\rho\|_*<1$.  Continuity of $\mathrm DH_\rho$ then makes $H_\rho$
a contraction in a sufficiently small ball around $p^*$ in this norm.
Its iterates converge to $p^*$ from this ball, proving local asymptotic
stability.  The spectral condition does not require a symmetric $K$;
for nonnormal matrices it need not imply contraction at every step in
the Euclidean norm.

For a possibly complex eigenvalue $\zeta$,
\begin{equation}
  |1-\rho+\rho\zeta|^2
  =1+2\rho(\operatorname{Re}\zeta-1)+\rho^2|1-\zeta|^2.
  \label{eq:complex-feedback-stability}
\end{equation}
Hence a mode is strictly damped exactly when
$\operatorname{Re}\zeta<1$ and
$0<\rho<2(1-\operatorname{Re}\zeta)/|1-\zeta|^2$, subject also to
$\rho\leq1$ for the relaxed update considered here.
This must hold for every mode to obtain the proposition's stability result.
For real $\zeta<-1$ the condition reduces to $0<\rho<2/(1-\zeta)$,
whereas direct replacement has amplification factor $\zeta$ and is unstable
in that direction.  For $0<\zeta<1$, the relaxed factor lies between
$\zeta$ and one, so relaxation slows convergence in this mode.
Real $\zeta>1$ remains unstable for every $\rho>0$.
\end{proof}

The scalar example $\zeta=-2$, $\rho=1/2$ illustrates this spectral
mechanism.  The proposition concerns a fixed-time local map; the sampler
uses a changing time and query, for which a uniform stability analysis would
require additional bounds on the sequence of linearizations.

\subsection{Finite-step residual control and accumulated error}
\label{app:discrete-residual}

\begin{proof}[Proof of Proposition~\ref{prop:discrete-residual}]
For $a^{(k+1)}=a^{(k)}+hu^{(k)}$, the fundamental theorem of calculus
gives $c^{(k+1)}=c^{(k)}+hJ^{(k)}u^{(k)}+R^{(k)}$, where
\begin{equation}
  R^{(k)}=h\int_0^1
  [J_{\C}(a^{(k)}+rhu^{(k)})-J^{(k)}]u^{(k)}\,\dd r,
  \quad \|R^{(k)}\|\leq\tfrac12 L_Jh^2\|u^{(k)}\|^2.
\end{equation}
The exact unregularized solve gives
$J^{(k)}u^{(k)}=-\kappa_kc^{(k)}$, proving
Eq.~\eqref{eq:discrete-residual}.  For an affine constraint the Jacobian is
constant, so $R^{(k)}=0$.  The factor $1-h\kappa_k$ has magnitude less
than one precisely when $0<h\kappa_k<2$, and is nonnegative when
$0\leq h\kappa_k\leq1$.
\end{proof}

For a full-row-rank affine constraint $\C(a)=Ha-y$, exact unregularized
normal correction can also be written as a split step:
\begin{equation}
\begin{aligned}
a^{(k+1)}={}&a^{(k)}+\rho_kP_T(d^{(k)}-a^{(k)})\\
&-h\eta(t_k)t_kP_Tq^{(k)}\\
&-\alpha_kH^\top(HH^\top)^{-1}\C(a^{(k)}).
\end{aligned}
\label{eq:affine-split-step}
\end{equation}
Since $HP_T=0$, this step has the exact affine residual law
\begin{equation}
  Ha^{(k+1)}-y=(1-\alpha_k)(Ha^{(k)}-y).
  \label{eq:affine-residual-law}
\end{equation}
The explicit Euler update in Algorithm~\ref{alg:persistent-mafm} uses
$\alpha_k=h\kappa_k$.  With frozen $\kappa_k$, the alternatives
$\alpha_k=1-e^{-h\kappa_k}$ and $\alpha_k=1$ give exponential residual
decay and one-step affine feasibility, respectively.  The split form also
applies when $\rho_k=0$.

For a regularized or inexact solve, let
$G^{(k)}=J^{(k)}(J^{(k)})^\top$ and define the
algebraic residual
$r_\lambda^{(k)}=(G^{(k)}+\delta I_m)\widehat\lambda^{(k)}
+J^{(k)}w^{(k)}+\kappa_kc^{(k)}$.
With $u^{(k)}=w^{(k)}+(J^{(k)})^\top\widehat\lambda^{(k)}$,
direct substitution gives
\begin{align}
  c^{(k+1)}&=(1-h\kappa_k)c^{(k)}+he^{(k)}+R^{(k)},\\
  e^{(k)}&=\delta(G^{(k)}+\delta I_m)^{-1}
       (J^{(k)}w^{(k)}+\kappa_kc^{(k)})\\
       &\quad+G^{(k)}(G^{(k)}+\delta I_m)^{-1}r_\lambda^{(k)}.
  \label{eq:discrete-solve-forcing}
\end{align}
Here the inverse exists for $\delta>0$, or for $\delta=0$ with full-row-rank
$J^{(k)}$.  The two terms in $e^{(k)}$ separate regularization from
linear-solve error.  If the smallest eigenvalue of $G^{(k)}$ is
$\ell_k>0$, then
$\|e^{(k)}\|\leq[\delta/(\ell_k+\delta)]
\|J^{(k)}w^{(k)}+\kappa_kc^{(k)}\|+\|r_\lambda^{(k)}\|$.

Set $q_k=|1-h\kappa_k|$ and
$f_k=h\|e^{(k)}\|+\tfrac12 L_Jh^2\|u^{(k)}\|^2$, using a common Lipschitz bound
over the step segments.  Iteration yields
\begin{equation}
  \|c^{(N)}\|\leq\left(\prod_{k=0}^{N-1}q_k\right)\|c^{(0)}\|
  +\sum_{i=0}^{N-1}f_i\prod_{j=i+1}^{N-1}q_j.
  \label{eq:accumulated-residual}
\end{equation}
In particular, for constant $\kappa>0$, $0<h\kappa\leq1$,
$\|u^{(k)}\|\leq U$, and $\|e^{(k)}\|\leq\varepsilon_e$, this becomes
\begin{equation}
  \|c^{(N)}\|\leq(1-h\kappa)^N\|c^{(0)}\|
  +\frac{\varepsilon_e+\tfrac12 L_JhU^2}{\kappa}
   [1-(1-h\kappa)^N].
  \label{eq:residual-error-floor}
\end{equation}
Under these uniform bounds, the estimate quantifies both the contraction of
the initial residual and the accumulated effects of curvature and numerical
corrections for the explicit anchor step in Algorithm~\ref{alg:persistent-mafm}.

\subsection{Endpoint stability on a common query region}
\label{app:endpoint-stability}

\begin{proposition}[Local model-perturbation bound]
\label{prop:endpoint-stability}
Let $a_t^*$ use a reference endpoint field $D^*$ and its input adjoint
$g^*$, and let $a_t^\theta$ use $D_\theta$ and $g^\theta$.  Couple the paths
through the same source, constraint, and schedules, with $\delta=0$, exact
multiplier solves, and full-row-rank $J_{\C}$ along a common trajectory
neighborhood.  Suppose the anchor vector field has an integrable
$L(t)$-Lipschitz bound there and, at a common state in $\Omega_t$,
$\|D_\theta-D^*\|\leq\varepsilon_D(t)$ and
$\|g^\theta-g^*\|\leq\varepsilon_g(t)$, with finite error
integrals.  Then
\begin{equation}
  \|a_t^\theta-a_t^*\|
  \leq e^{\int_0^tL(q)\dd q}
  \left[\|a_0^\theta-a_0^*\|
  +\int_0^t\left(
    \gamma(r)\varepsilon_D(r)
    +\eta(r)r\varepsilon_g(r)
  \right)\dd r\right].
  \label{eq:endpoint-bound}
\end{equation}
\end{proposition}

\begin{proof}
Let $V_{D,g}(a,t;x_0)$ denote the exact unregularized anchor vector field
after inserting Eq.~\eqref{eq:query-state}.  At a common anchor,
\begin{equation}
\|V_{D_\theta,g^\theta}(a,t;x_0)-V_{D^*,g^*}(a,t;x_0)\|
\leq\gamma(t)\varepsilon_D(t)+\eta(t)t\varepsilon_g(t),
\end{equation}
because the orthogonal projector $P_T(a)$ has norm at most one.  Comparing
the two paths and using the anchor-Lipschitz bound gives
\begin{equation}
  \|V_{D_\theta,g^\theta}(a_t^\theta,t;x_0)
      -V_{D^*,g^*}(a_t^*,t;x_0)\|
  \leq L(t)\|a_t^\theta-a_t^*\|
  +\gamma(t)\varepsilon_D(t)+\eta(t)t\varepsilon_g(t).
\end{equation}
Gr\"onwall's inequality gives Eq.~\eqref{eq:endpoint-bound}.
\end{proof}

For exact unregularized solves, Eq.~\eqref{eq:projected-endpoint-error}
also permits replacing $\varepsilon_D(t)$ in the trajectory bound by a
uniform bound on
$\|P_T(a)[D_\theta(x,t)-D^*(x,t)]\|$ over the same queried region.
For $\eta\equiv0$, the adjoint-error term vanishes.  If the query region
extends beyond the training-path support, the reference field $D^*$ requires
a specified extension with the assumed regularity.  The optional target-law
comparison in Eq.~\eqref{eq:target-law-bound} addresses a separate question
from this pathwise model-perturbation bound.

\section{Additional algorithmic details}
\label{app:algorithms}

\subsection{Optional adjoint-step clipping}
\label{app:trust-clipping}
The default $\operatorname{clip}_k$ in Algorithm~\ref{alg:persistent-mafm}
is the identity.  For full-row-rank $J_{\C}(a^{(k)})$, an optional
radius $\tau_k$ and norm $\|\cdot\|_\star$ define
\begin{equation}
  \operatorname{clip}_k(v)=
  \min\left\{1,\frac{\tau_k}
  {h\eta(t_k)t_k\|P_T(a^{(k)})v\|_\star+\varepsilon}\right\}v,
  \qquad \varepsilon>0.
  \label{eq:optional-tangent-clipping}
\end{equation}
Taking $\tau_k=\infty$ recovers the default.  A finite radius bounds the
projected adjoint displacement in $\|\cdot\|_\star$; the VJP is skipped
when $\eta(t_k)t_k=0$.

\subsection{Poisson adjoint scaling}
\label{app:poisson-scaling}
The main text uses the ordinary residual norm and input gradient.  The
reported partially observed Poisson sampler uses fixed channel scales
$s_U,s_F>0$ from the training data.  With $m$ observed $U$ entries and
$\C(z)=Hz-y$, its endpoint energy and physical guidance direction are
\begin{equation}
  \mathcal E_t^{\rm P}(x)
  =\frac{\|HD_\theta(x,t)-y\|^2}{2m s_U^2},
  \qquad
  g_t^{\rm P}
  =\bigl(s_U^2[\nabla_x\mathcal E_t^{\rm P}]_U,
          s_F^2[\nabla_x\mathcal E_t^{\rm P}]_F\bigr).
  \label{eq:poisson-guidance-scale}
\end{equation}
Its discrete adjoint step uses
$q^{(k)}=\operatorname{clip}_k(g_{t_k}^{\rm P})$, with the reported
per-step guidance coefficient corresponding to $h\eta(t_k)$ in the split
update of Eq.~\eqref{eq:affine-split-step}.  The channel factors adjust the
relative $U$ and $F$ motion; they are part of this numerical
specialization, not needed for the core dynamics.  The multiplier still
enforces the unscaled observation equation $Ha=y$.  The projected-adjoint
identity remains valid with $g_t^{\rm P}$ substituted for $g_t$.
The Poisson clipping norm is the root mean square of componentwise $v/s$,
where $s$ equals $s_U$ or $s_F$ by channel; its radius is the configured
trust factor times the norm of the tangent tracking displacement
$P_T(a^{(k)})(d^{(k)}-a^{(k)})$.

\subsection{\clarrev{How the anchor correction reaches the next query}}
\begin{revision}[clarityink]
In Fig.~\ref{fig:pipeline}, the frozen endpoint model takes the current query
$x^{(k)}$ and predicts $d^{(k)}$. The sampler keeps $a^{(k)}$ across steps
as its current candidate for the physical endpoint. The multiplier acts
on this anchor. For a full-row-rank constraint Jacobian $J_k$ at the
anchor and an exact unregularized solve, its velocity correction is
$-J_k^\dagger[J_kw^{(k)}+\kappa(t_k)\C(a^{(k)})]$, where
$J_k^\dagger=J_k^\top(J_kJ_k^\top)^{-1}$. The input VJP in
Eq.~\eqref{eq:input-adjoint} instead differentiates the loss of the
predicted endpoint through the frozen model to build the candidate velocity.

For an affine constraint, the tangential update preserves the current
constraint value and the normal update contracts its residual. For example,
if a Burgers frame has mean $0.8$ and the target is $0.9$, a normal step
that halves the residual moves the mean to $0.85$. The tangential update
can use the new endpoint prediction to change the waveform while
preserving the frame mean. At fixed $x_0$ and $t_{k+1}$, an anchor
displacement $\Delta a$ changes the next query by $t_{k+1}\Delta a$.
Intermediate queries can therefore remain outside the feasible set.
At $t=1$ the query is the final anchor and has the same constraint residual.
\end{revision}

\input{sections/appendix_protocols}

\FloatBarrier
\input{sections/appendix_experiments}

\clearpage
\section{Scope and practical limitations}
\label{app:numerical-boundaries}

The residual-contraction law describes ideal continuous dynamics with a
full-rank constraint Jacobian and exact multiplier solves. In a numerical
sampler, Gram regularization, solve error, and time discretization can alter
that law. A finite attraction budget $\int_0^1\kappa(t)\dd t$ also gives a
finite reduction unless the sampler starts feasible or applies an exact normal
step.
Thus a small measured constraint error supports numerical feasibility, while
matching a separately prescribed law on the feasible intersection remains a
distinct question (Eq.~\eqref{eq:target-law-bound}).

The quality of the guidance depends on the frozen endpoint model. Its
conditional-mean prediction may average modes or become less reliable away
from the training distribution; the $\eta>0$ adjoint is useful when the
predicted-endpoint residual and input sensitivity convey a stable correction
(Appendix~\ref{app:adjoint-effect}). Each active adjoint step adds one input
VJP, so its benefit must be read together with the sampling-time comparison
in Table~\ref{tab:compute-cost}. The model parameters remain frozen.

The empirical evidence is limited to one training seed per task.
\rev{The main Euler and controlled-NS comparisons use the training energy.
Appendix~\ref{app:shifted-energy} tests the same checkpoints at $0.8$,
$1.2$, and $2$ times the training energy. The shared terminal projection brings
MAFM within the reported energy tolerance at the two moderate shifts.
At twice the training energy, projected MAFM still meets the tolerance,
but the PDE relative $L_2$ errors rise to $9.85\%$ on Euler and $10.01\%$ on
controlled NS. These results cover the three tested targets and one checkpoint
per system.}
The Burgers samples show mild fluctuation-amplitude
contraction, and reaction--diffusion samples retain a small negative-value
tail. These observations define the scope of the present results and motivate
broader seed and condition studies.

%% file: sections/appendix_protocols.tex
\section{Experimental protocols}
\label{app:protocols}
\label{app:task-constraints}

This section specifies the physical data, equality conditions, and evaluation
for the six examples in Section~\ref{sec:experiments}. In every case, the
unconditional model and each constrained sampler act on the same complete
physical object. The numerical PDE solvers generate reference data and measure
physical consistency; they do not supply an unobserved field to the sampler.
Table~\ref{tab:case-overview} separates structure learned from training data
from information explicitly supplied at inference. Here $\tau$ denotes
physical time, while $t$ denotes the flow or sampling time.
The notation $\langle f\rangle_\Omega$ means the spatial average, evaluated
as the corresponding grid average, and $\xi$ denotes the Burgers spatial
coordinate. The Euler and controlled-NS entries describe their fixed-energy,
energy-only tests.

\input{tables/case_overview}

\paragraph{Shared comparison and metrics.}
Methods within a case use the same frozen unconditional model, paired Gaussian
sources, prescribed conditions, and nominal number of function evaluations
(NFE). A constraint error (CE) measures the residual of the condition supplied
at sampling. It is distinct from a physical-consistency error: for PDE
trajectories, we evolve each generated initial state with an independent
numerical solver and compare its later frames with the generated trajectory
(own-IC relative $L_2$); for Poisson, we solve from the generated $F$ and
compare with the generated $U$. Distributional and paired-truth diagnostics
are reported separately. The Poisson protocol requires both mean and q99
observation CE below $10^{-6}$. For concise comparisons, we mark mass CE below
$10^{-6}$ and fixed-energy error below $10^{-5}$ as numerically feasible;
these two bands are reporting conventions. The sphere's $10^{-3}$ plane
tolerance belongs only to its joint-support diagnostic, not to a universal
feasibility rule. Once a method reaches the applicable numerical band, the
physical and distributional metrics distinguish its samples.

\subsection{Sphere--plane intersection}
\label{app:sphere-protocol}
The training law and the inference-time plane condition are
\begin{equation}
X\sim\operatorname{Unif}(\mathbb S^2)\subset\mathbb R^3,
\qquad X_3=0.
\end{equation}
We draw a
standard Gaussian vector and normalize it to unit length, then train an
unconditional velocity MLP on full Cartesian three-vectors with a Gaussian
source and linear flow-matching bridge. The width-192, depth-three MLP uses
velocity MSE, batch size 512, and 100,000 optimizer updates for its frozen
checkpoint. It receives no sphere equation or plane condition. At inference
the only supplied condition is the plane equation above;
its intersection with the learned support is the unit circle in that plane.
There is no physical-time evolution or boundary condition in this algebraic
example. The paired 10,000-sample comparison uses 100 NFE and independent
uniform-circle references. We report plane residual, radial and angular
distribution diagnostics, sliced Wasserstein distance (SWD), and joint
support coverage over several tolerances. The radius is an evaluation
diagnostic, not an additional constraint given to the sampler.

\subsection{Partially observed Poisson equation}
\label{app:poisson-protocol}
On $\Omega=(0,1)^2$, the data satisfy
\begin{equation}
\begin{aligned}
-\Delta U&=F &&\text{in }\Omega,\\
U&=0 &&\text{on }\partial\Omega.
\end{aligned}
\end{equation}
Each physical state is the joint pair $(U,F)$ on
the $64\times64$ interior grid ($h=1/65$); boundary nodes are fixed rather
than generated. To draw $F$, independent grid Gaussian noise is smoothed by
a reflected Gaussian kernel of width three grid cells, scaled to spatial RMS
about 20, and given an independent Gaussian constant offset with standard
deviation 5. A float64 five-point Dirichlet solve supplies $U$. The joint
physical-field network predicts both channels and minimizes a channel-scaled
full-field endpoint MSE. It uses 8,192 training and 1,024 validation pairs;
the width-32, depth-three joint U-Net trains with batch size 32. The final
40,000-update AdamW run selects its checkpoint by validation loss. Fixed
channel scales balance the two physical units and do not change
the generated state space.

At inference, the supplied observation equation is
\begin{equation}
H_UU=y_{\mathrm{obs}},
\end{equation}
which selects either 64 or 256 distinct interior $U$ values;
$F$ is never observed. Two fixed sensor layouts are evaluated for each count,
with 64 independently generated conditions and eight paired draws per
condition, at 100 NFE. The held-out pair is used only to form observations and
evaluation targets. Observation CE is normalized by the truth-$U$ RMS; PDE
consistency is $\|\widehat U-S_h(\widehat F)\|_2/\|S_h(\widehat F)\|_2$,
where $S_h$ is the same discrete boundary-value solver. Errors against the
unobserved $U$ and $F$ are supplementary paired-truth diagnostics, not
conditional-distribution distances. The Poisson-specific adjoint scaling and
tangent-step radius are defined in Appendix~\ref{app:poisson-scaling}.

\subsection{Periodic viscous Burgers equation}
\label{app:burgers-protocol}
We solve
\begin{equation}
\partial_\tau u+\partial_\xi\!\left(\frac{u^2}{2}\right)
=0.01\partial_{\xi\xi}u
\end{equation}
on
$\xi\in[0,1)$ with periodic boundary conditions, recording 65 frames over
$\tau\in[0,1]$ on 512 grid points. Each initial field is
$u_0=m+A g$: $m\sim\mathrm{Unif}[0.3,1.3]$,
$A\sim\mathrm{Unif}[0.3,0.8]$, and $g$ is a zero-mean, unit-RMS smooth
periodic Gaussian field built from sine and cosine frequencies 1--8 with
amplitude proportional to $(1+k^2)^{-1}$. A float64 Cole--Hopf solve generates
the physical trajectories. Its Fourier calculations are numerical solver
operations; the jointly trained endpoint network sees complete
$[65,1,512]$ trajectories. Training uses 4,096 trajectories and 512
validation trajectories. A joint circular-convolution U-Net mixes all frames
as channels and uses a full-trajectory endpoint MSE with batch size 32. The
reported model is the selected 52,000-update checkpoint after the registered
time-allocation continuation.

The inference condition prescribes the spatial mean $m_\star$ of every
generated frame. Held-out tests use $m_\star\in\{0.55,0.90,1.05\}$, with 256
independent sources at each value and 100 NFE. For each raw generated initial
field, an independent Burgers solve produces the own-IC future reference.
We compare the remaining 64 frames by relative $L_2$; mean-subtracted
fluctuation error and distributional statistics check whether a low error
merely reflects correct means. Mean CE is the framewise RMS deviation from
$m_\star$, normalized by $|m_\star|$.

\rev{At each prescribed mean we independently generate 256 reference initial fields with that mean and solve Burgers. Field SWD compares the generated and reference full trajectories separately at each mean, using the same 64 fixed slicing directions. The aggregate is the average over the three means. This reference ensemble is separate from the solver rollout initialized from each generated sample.}

\subsection{Mass-conserving reaction--diffusion system}
\label{app:rd-protocol}
On $\Omega=[0,10]^2$ with no-flux boundaries, the two concentrations obey
\begin{equation}
\begin{aligned}
\partial_\tau U&=0.1\Delta U+\mathcal R(U,V),&
\partial_\tau V&=10\Delta V-\mathcal R(U,V),\\
\mathcal R(U,V)&=\left(0.067+\frac{U^2}{1+U^2}\right)V-U.
\end{aligned}
\end{equation}
The opposite reaction signs and no-flux boundary conserve the spatial mean
of $U+V$. Initial states are positive: a smooth random total concentration
is normalized to a mass drawn uniformly from $[2,2.5]$, then split into
the two species by a smooth random fraction with one to three reflected
Gaussian bumps. A conservative cell-centered Neumann solver uses diffusion
substeps and Runge--Kutta reaction steps to produce 33 frames over
$\tau\in[0,40]$ on a native $64\times64$ grid. The joint endpoint U-Net
learns complete $[33,2,64,64]$ trajectories from 4,096 training examples,
with 512 validation examples and 60,000 optimizer updates. Its width is 48,
its batch size is eight, and full-field endpoint MSE trains both species and
all frames together; validation loss selects the frozen checkpoint.

The sampler is told only the desired total mass at every frame. The held-out
values are $2.15$, $2.25$, and $2.35$, each with 128 paired sources at 100
NFE. CE is the framewise residual of the mean of $U+V$; an independent
rollout from each generated first frame gives own-IC relative $L_2$ over the
remaining 32 frames. Species-wise errors, distributional distances, and the
fraction of negative generated concentrations are supplementary checks.
Negative values are counted on the raw output rather than clipped.

\subsection{Two-dimensional Euler flow}
\label{app:euler-protocol}
The data solve
\begin{equation}
\begin{aligned}
\partial_\tau\boldsymbol u
+(\boldsymbol u\cdot\nabla)\boldsymbol u+\nabla p&=0,\\
\nabla\cdot\boldsymbol u&=0
\end{aligned}
\end{equation}
on the periodic torus $[0,2\pi)^2$.
Smooth random velocity fields combine randomly phased low-frequency patterns
and smoothed noise; they are projected to zero-mean, divergence-free fields
and normalized to the common kinetic-energy level before numerical rollout.
The resulting training objects are nine complete two-component velocity
frames on a $49\times49$ grid over $\tau\in[0,1]$. A physical-field endpoint
network learns their joint distribution using full-field MSE and an auxiliary
full-field spectral error. It has width 48, four spectral blocks, and 12
modes per spatial direction; these features are internal computations, not
generated states. Training has 4,096 histories, 1,024 validation histories,
batch size 16, and 48,000 updates.

The energy-only inference test supplies the target kinetic energy at every
frame; divergence-free and zero-mean structure must still come from the
learned law. \rev{Guidance strengths for the original fixed-energy comparisons
were selected on a disjoint 64-sample validation set.} On 512 held-out paired
sources, we measure energy
error, divergence, spatial mean, field and feature distribution distances,
and own-IC rollout error from each generated first frame.
\rev{In Table~\ref{tab:dynamics-main}(c), the forward budget is 100 per sample for
most methods. Stop-gradient guidance uses 199, LDF uses 198, and D-Flow uses
4{,}100 including recomputation (Appendix~\ref{app:additional-inverse-baselines}).}
\rev{Before the reference solve, the generated first frame is projected to a divergence-free, zero-mean field and normalized to the specified energy. The rollout error compares the reference with the unmodified generated future frames. For energy CE, we take the maximum relative energy deviation over frames for each sample and then average over samples.} The separate
joint-oracle comparison supplies divergence and zero mean as additional
conditions and is reported in Table~\ref{tab:oracle-full}; it does not enter
the energy-only ranking.

\subsection{Energy-controlled Navier--Stokes flow}
\label{app:ns-protocol}
The controlled system uses the same periodic domain, grid, nine-frame
horizon, and initial-field family as Euler, with
\begin{equation}
\partial_\tau\boldsymbol u=
-\mathsf P[(\boldsymbol u\cdot\nabla)\boldsymbol u]
+0.02\Delta\boldsymbol u+\beta_{\rm NS}(\tau)\boldsymbol u,
\qquad
\beta_{\rm NS}(\tau)=
-\frac{\langle\boldsymbol u,
-\mathsf P[(\boldsymbol u\cdot\nabla)\boldsymbol u]
+0.02\Delta\boldsymbol u\rangle}
{\langle\boldsymbol u,\boldsymbol u\rangle}.
\end{equation}
Here $\mathsf P$ is the divergence-free, zero-mean Leray projection; the
feedback offsets instantaneous energy change. The solver constructs complete
Cartesian velocity histories, and the same physical-field architecture and
training sizes as Euler learn them jointly. At inference, only framewise
energy is imposed. The fixed-energy 512-sample held-out comparison uses
the same forward budgets and constraint, distributional, and own-IC diagnostics
as Euler. \rev{The controlled-NS reference additionally bandlimits the initial field. The original fixed-energy tests use the training energy; Appendix~\ref{app:shifted-energy} reports the separate shifted-target controls.}

%% file: tables/case_overview.tex
\begin{table}[t]
\centering
\caption{Equations, learned structure, and supplied conditions for the six benchmarks.}
\label{tab:task-constraint-interfaces}
\label{tab:case-overview}
\footnotesize
\setlength{\tabcolsep}{2.5pt}
\renewcommand{\arraystretch}{1.16}
\begin{tabularx}{\linewidth}{@{}
  >{\raggedright\arraybackslash}p{0.115\linewidth}
  >{\raggedright\arraybackslash}p{0.255\linewidth}
  >{\raggedright\arraybackslash}X
  >{\raggedright\arraybackslash}X
  >{\raggedright\arraybackslash}p{0.14\linewidth}@{}}
\toprule
Benchmark & Equation or data distribution
& Structure learned from data
& Explicit inference-time condition
& Constraint type \\
\midrule
Sphere--plane
& $X\sim\operatorname{Unif}(\mathbb S^2)$
& Uniform unit-sphere distribution, $\|X\|_2=1$.
& Plane $X_3=0$; the desired support is the intersection circle.
& Affine, single-coordinate algebraic \\
\addlinespace[3pt]
Poisson
& $-\Delta U=F$,\newline $U|_{\partial\Omega}=0$
& Joint $(U,F)$ samples satisfy the discrete Poisson equation and homogeneous boundary conditions.
& $H_UU=y_{\mathrm{obs}}$; only $U$ is observed, with no direct observations of $F$.
& Sparse, local linear observations \\
\addlinespace[3pt]
Burgers
& $\partial_\tau u+\partial_\xi(u^2/2)$\newline $\qquad=0.01\partial_{\xi\xi}u$
& Periodic viscous dynamics and their trajectory statistics.
& Prescribed spatial mean $\langle u(\tau)\rangle_\Omega=\mu_\star$ at every frame.
& Framewise global linear integral \\
\addlinespace[3pt]
Reaction--diffusion
& $\partial_\tau U=0.1\Delta U+\mathcal R$,\newline
  $\partial_\tau V=10\Delta V-\mathcal R$
& No-flux nonlinear dynamics, positive concentration support, and trajectory statistics.
& Prescribed total mass $\langle U(\tau)+V(\tau)\rangle_\Omega=\mu_\star$ at every frame.
& Framewise global linear integral \\
\addlinespace[3pt]
Euler
& $\partial_\tau\boldsymbol u+$\newline
  $(\boldsymbol u\cdot\nabla)\boldsymbol u+\nabla p=0$
& Incompressible, zero-mean velocity trajectories generated by Euler dynamics.
& Prescribed kinetic energy at every frame; incompressibility and zero mean are left to the learned prior.
& Global nonlinear quadratic \\
\addlinespace[3pt]
Controlled NS
& $\partial_\tau\boldsymbol u=-\mathsf P[(\boldsymbol u\cdot\nabla)\boldsymbol u]$\newline
  $\quad+\nu\Delta\boldsymbol u+\beta_{\mathrm{NS}}(\tau)\boldsymbol u$
& Incompressible, zero-mean velocity trajectories generated by controlled NS dynamics.
& Prescribed kinetic energy at every frame; incompressibility and zero mean are left to the learned prior.
& Global nonlinear quadratic \\
\bottomrule
\end{tabularx}
\end{table}

%% file: sections/appendix_experiments.tex
\section{Supplementary results and diagnostics}
\label{app:complete-results}
\label{app:experiments}

The tables below give the extended metrics behind the compact main-text
comparison. They include the 256-sensor Poisson setting, matched-model
ablations, and the separate three-constraint flow comparison. The subsequent
diagnostics and sample panels clarify particular behaviors without changing
the evaluation conditions in Appendix~\ref{app:protocols}.

\paragraph{Main-figure visualization protocol.}
In Figure~\ref{fig:joint-main}(a), circles mark the shared observed sites;
the columns show generated fields, sensor mismatch, and
$|\widehat U-S_h(\widehat F)|$. The numerical solver $S_h$ is used only for
evaluation, and no hidden paired field is supplied as a visual target. The
illustrated PCFM row uses PCFM-Relaxed in the Poisson panel and PCFM-Base in
the controlled-NS panel, the registered variants for those tasks. Fields and
residual maps show individual samples; the tables report full evaluation sets.

\subsection{Extended metric tables}

Table~\ref{tab:sphere-full}'s plane indicator uses the $10^{-3}$ tolerance
inside Joint coverage. In Table~\ref{tab:poisson-full}, CE is the mean
observation residual and PDE is the relative discrepancy between the
generated $U$ and the numerical solution from its generated $F$; two fixed
sensor layouts contribute to each density. In the Burgers table CE is the
mean-trajectory mass residual. Burgers and reaction--diffusion relative
$L_2$ entries are percentages, including their fluctuation and q95 columns.

\input{tables/sphere_full}
\input{tables/poisson_full}
\input{tables/conservation_full}
\FloatBarrier
\input{tables/core_ablation_full}

The tables report values to at most four significant digits.
Table~\ref{tab:poisson-full} adds the 256-sensor comparison.
Table~\ref{tab:core-ablation-full} reports 10,000-resample paired intervals;
Poisson resamples conditions while retaining layouts and draws, and the
dynamical tasks resample sample IDs within each prescribed-mass stratum.
Its $\Delta$ column is ablation minus Full MAFM: Poisson uses PDE-consistency
relative $L_2$, while the dynamical cases use own-IC relative $L_2$. Field
SWD was not part of this Poisson ablation comparison.

\subsection{Three-constraint oracle comparison}

\rev{Table~\ref{tab:dynamics-main}(c) reports the 512-sample energy-only
comparison, including D-Flow and PnP-Flow. Their protocols and extended
results are in Appendix~\ref{app:additional-inverse-baselines}.}
The corresponding joint-oracle comparison below supplies energy, divergence,
and zero mean simultaneously. It records projection behavior and the effect
of additional analytic information.

\input{tables/oracle_full}

\subsection{Additional diagnostics}
\label{app:diagnostics}

\paragraph{Matched ablations.}
The variants in Table~\ref{tab:core-ablation-full} share each case's frozen
checkpoint, paired sources, conditions, and 100 NFE. Each own-IC reference is
solved from the initial condition produced by that variant. This isolates the
inference mechanism from changes in model training or reference selection.

\paragraph{Terminal projection.}
An optional final projection sharpens numerical feasibility without explaining
the consistency advantage: sphere SWD and the Burgers and reaction--diffusion
own-IC errors barely change. In the 512-sample flow tests, mean energy error
falls from about $6.9\times10^{-6}$ to about $2.4\times10^{-7}$ for both
Euler and controlled NS, while own-IC error and field SWD are effectively
unchanged.

\begin{figure}[!t]
\centering
\includegraphics[width=\linewidth]{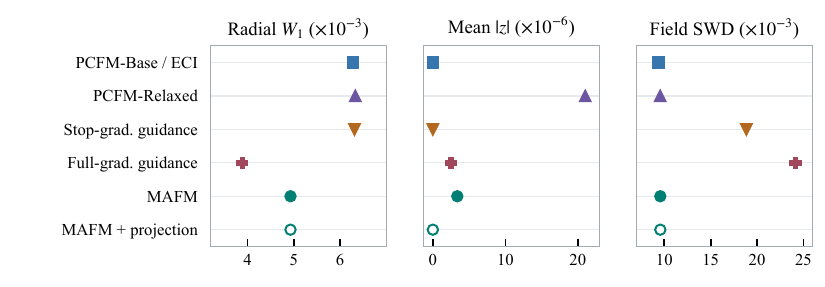}
\caption{\rev{Sphere--plane diagnostics. Each column has its own scale.
Lower values are better on all three metrics.}}
\label{fig:sphere-metrics}
\par\medskip
\includegraphics[width=\linewidth]{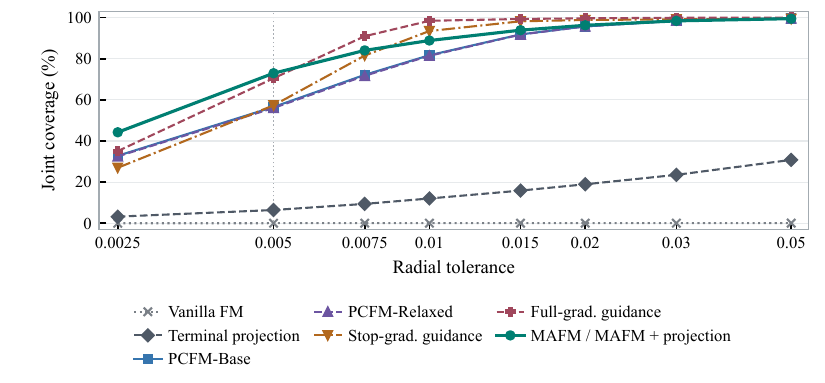}
\caption{\rev{Sphere--plane joint coverage at varying radial tolerances and
fixed plane tolerance $|z|\le 10^{-3}$. The MAFM curves with and without
terminal projection coincide at all reported tolerances.}}
\label{fig:sphere-threshold}
\end{figure}

\paragraph{Sphere threshold sensitivity.}
\rev{Figure~\ref{fig:sphere-metrics} compares radial $W_1$, mean plane
residual, and field SWD on separate axes. Full-gradient guidance attains a
smaller radial $W_1$ than MAFM and a higher field SWD. Terminal projection
reduces MAFM's mean plane residual to zero, with the same reported radial
$W_1$ and field SWD.} MAFM has the highest joint coverage at the strict
0.0025 and 0.005 support tolerances; conclusions at looser thresholds
are reported as sensitivity analysis rather than universal support claims.

\paragraph{Burgers uncertainty.}
The independent-sample bootstrap preserves MAFM's advantage
in own-IC rollout error. The analysis also exposes a remaining failure mode:
generated fluctuation amplitude is slightly contracted even when the mean
trajectory is exact.

\paragraph{Reaction--diffusion nonnegativity.}
Nonnegativity is not supplied to any sampler. MAFM's negative-value fraction is
$2.33\times10^{-4}$: better than PCFM variants but worse than Vanilla and
stop-gradient guidance. This auxiliary diagnostic does not reverse the primary
ranking and remains a target for future work.

\paragraph{PCFM normalization.}
Changing the relaxed-PCFM normalization can improve selected distribution
distances, but the same changes can worsen feasibility and own-IC rollout error.
The paper therefore reports the official Eq.~6 arm and labels operator-equivalent
duplicates instead of counting them as independent evidence.

\subsection{Inference-time computational cost}
\label{app:compute-cost}

\begin{revision}[costink]
Reusing the initial endpoint prediction gives $N$ model forward calls for
$N$ steps in Algorithm~\ref{alg:persistent-mafm}. Each step that uses input
adjoint guidance adds one vector--Jacobian product (VJP). Storing the anchor
and interpolating it with the fixed source cost $O(d)$ memory and work,
respectively.

We focus on $m\ll d$, with $m$ scalar constraints and physical dimension
$d$. The MAFM multiplier solve uses the Jacobian at the anchor and the
right-hand side in Eq.~\eqref{eq:unified-multiplier}. The Gauss--Newton
projection in PCFM~\citep{utkarsh2026physics} uses the Jacobian and residual
at the predicted endpoint. Given the Jacobian, each dense correction takes
$O(dm^2)$ to form the Gram matrix and $O(m^3)$ to factor and solve the
$m\times m$ system. A gridwise PDE constraint can make $m$ comparable to $d$,
bringing the generic dense cost to $O(d^3)$. The cost of obtaining the
Jacobian is separate. Sensor selection in Poisson gives a diagonal Gram
matrix. The mass and energy experiments instead use framewise mean and
radial corrections, respectively. In the main tests, we impose the
task-specific constraints and evaluate PDE consistency with an independent
solver.
\end{revision}

All samplers in a given case reuse the same frozen physical-field model, so
the comparisons below concern inference rather than repeated model training.
Table~\ref{tab:compute-cost} transcribes recorded remote sampling times for
two representative tasks: one fixed 64-sensor Poisson layout with 64 conditions
and eight draws per condition, and the 384-sample reaction--diffusion test over
three prescribed masses. The timers cover sample generation and transfer to
host memory, but exclude independent PDE-reference solves and metric
aggregation. Time units are stated per row because the two tasks use different
physical states and batch protocols; their absolute times should not be
compared across tasks.

\begin{table}[H]
\centering
\small
\setlength{\tabcolsep}{3pt}
\begin{tabular*}{\linewidth}{@{\extracolsep{\fill}}llccccc@{}}
\toprule
Case & Sampler & NFE & Input VJPs & Sampling time & CE & Physical RelL2 \\
\midrule
Poisson & Vanilla FM & 100 & 0 & 6.615 s/512 & 1.366 & 1.128\% \\
(64 sensors) & PCFM-Relaxed & 100 & 0 & 15.77 s/512 & 0 & 10.26\% \\
& MAFM & 100 & 100 & 17.88 s/512 & $1.696\mathrm{e}{-7}$ & 1.710\% \\
\midrule
Reaction--diffusion & PCFM-Base & 100 & 0 & 0.04249 s/sample & $5.894\mathrm{e}{-8}$ & 3.030\% \\
& ECI & 100 & 0 & 0.04361 s/sample & $5.855\mathrm{e}{-8}$ & 2.962\% \\
& MAFM & 100 & 0 & 0.04375 s/sample & $1.134\mathrm{e}{-7}$ & 2.740\% \\
& Full-grad. guidance & 100 & 100 & 0.08584 s/sample & $1.695\mathrm{e}{-3}$ & 1.563\% \\
\bottomrule
\end{tabular*}
\caption{Recorded inference cost and physical consistency for two frozen-model
evaluations. Poisson values use one fixed 64-sensor layout (512 samples);
reaction--diffusion values cover 384 samples. CE is the supplied-constraint
error. Lower physical RelL2 is preferable among numerically feasible methods.}
\label{tab:compute-cost}
\end{table}

\costrev{The recorded Poisson implementation executes 100 input VJPs, including the first call where the guidance coefficient is zero.}
MAFM uses 100 forward calls
and takes 17.88 seconds for the 512 outputs, compared with 15.77 seconds
for PCFM-Relaxed. Both satisfy the observation tolerance, while MAFM reduces
PDE RelL2 from 10.26\% to 1.710\%. The unguided Vanilla sampler is faster,
but its observation CE is far outside the feasible band.

\begin{revision}
\paragraph{Paired Poisson timing.}
In a separate ablation on two 64-sensor layouts, both variants use the
same frozen model and paired inputs, with 512 samples per layout. With the
input adjoint off, MAFM uses 100 forward calls and zero input VJPs and takes
5.23\,s per layout on average. With it on, MAFM uses 100 forward calls and
100 input VJPs and takes 14.04\,s. The timing difference includes the full
input-adjoint update.
\end{revision}

On reaction--diffusion, base MAFM needs no input VJP and takes 0.04375 seconds
per sample, close to ECI's 0.04361 and PCFM-Base's 0.04249, while reducing
feasible own-IC RelL2 to 2.740\% from 2.962\% and 3.030\%, respectively.

These two evaluations show that MAFM can improve feasible physical
consistency with limited additional sampling time relative to constrained
baselines. The gain is especially pronounced on Poisson, where the adjoint
also increases cost. The records do not establish a uniform runtime or
memory advantage across all six tasks.

\costrev{In a separate energy-only archive with 512 samples per case on an RTX 4090, base MAFM took 7.25 and 7.28 seconds on Euler and controlled NS. PCFM-Base took 6.72 and 6.73 seconds, respectively. Both samplers use 100 forward calls and no input VJPs in these tests.
PCFM projects each predicted endpoint radially, while MAFM tracks the
endpoint tangentially and contracts the radial deviation of the persistent
anchor. The recorded times cover that batch and exclude PDE-reference
evaluation.}

\input{sections/rebuttal_results}

\subsection{Qualitative samples}
\label{app:samples}

These panels illustrate individual outputs from the six cases. The sphere
panel checks recovery of the circular intersection from a supplied plane and
a learned sphere prior; the radius was never imposed during sampling.
The Poisson $U$ and $F$ panels show corresponding joint-field samples.
The trajectory and vorticity panels show selected outputs rather than the
full evaluation set, so quantitative conclusions follow from
Figure~\ref{fig:sphere-metrics}, Table~\ref{tab:sphere-full}, and the threshold
sensitivity in Figure~\ref{fig:sphere-threshold}.

\begin{figure}[H]
\centering
\includegraphics[width=.58\linewidth]{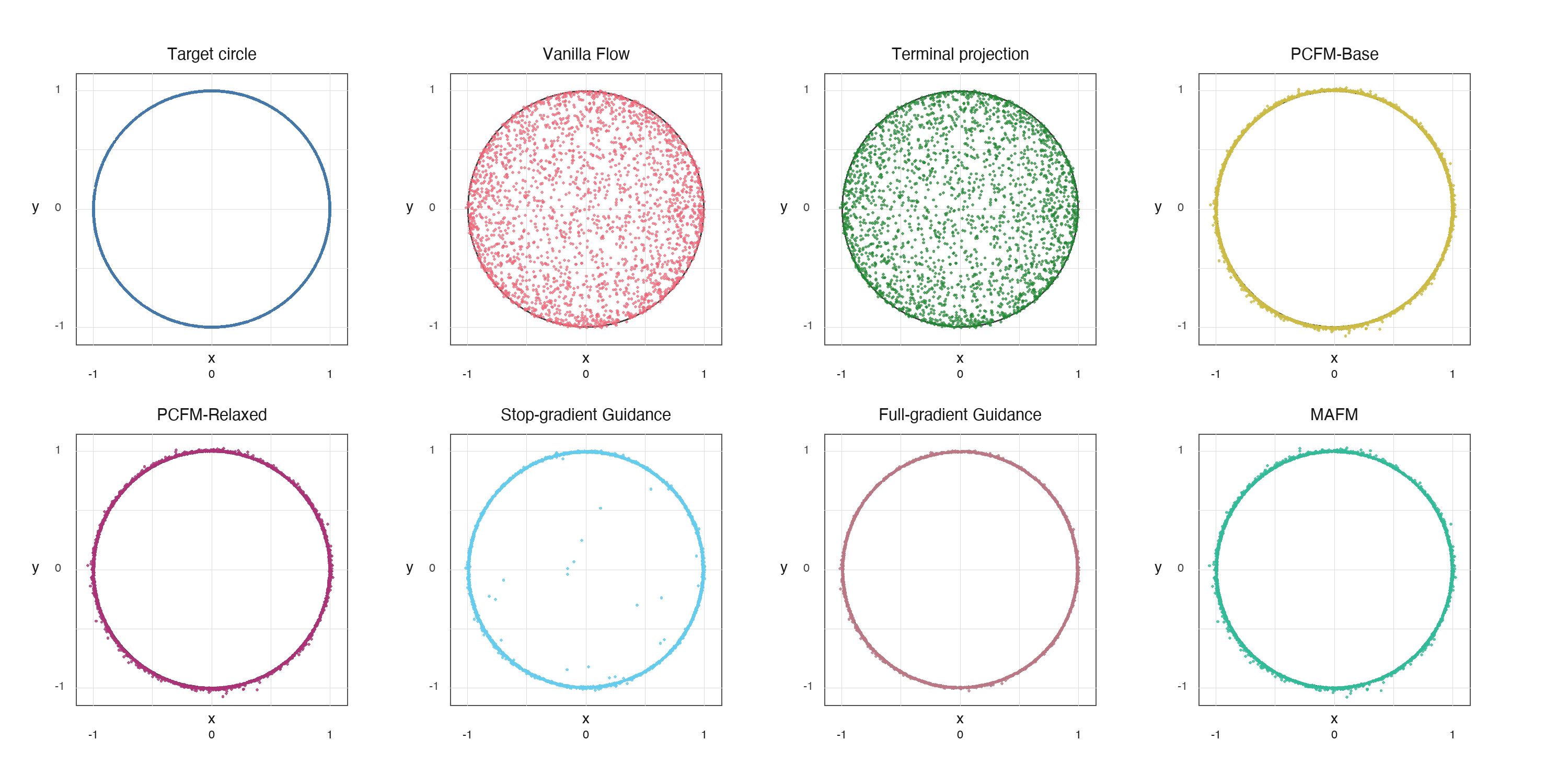}
\caption{Sphere--plane samples.}
\label{fig:sphere-samples}
\end{figure}

\begin{figure}[!t]
\centering
\includegraphics[trim=0 0 765.5bp 38bp,clip,width=.66\linewidth]{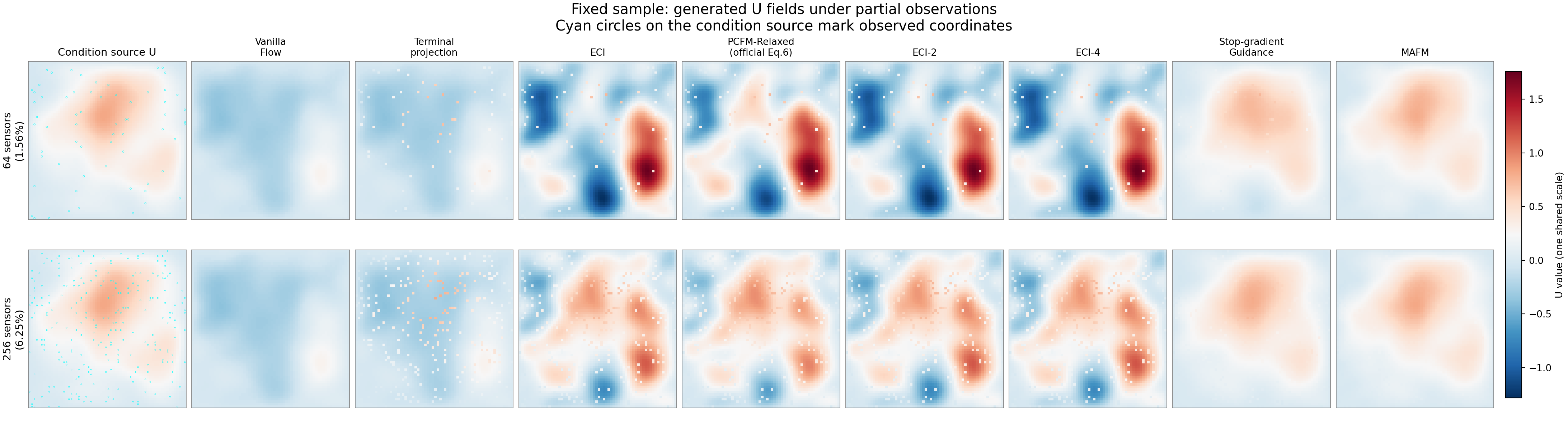}%
\hfill
\includegraphics[trim=1235.4bp 0 0 38bp,clip,width=.315\linewidth]{figures/poisson_samples_u.png}
\caption{Poisson generated solution fields $U$ under partial observations. One
representative ECI column is shown; the ECI-2 and ECI-4 variants remain in
Table~\ref{tab:poisson-full}.}
\label{fig:poisson-u-samples}
\end{figure}

\begin{figure}[!t]
\centering
\includegraphics[trim=0 0 765.5bp 20bp,clip,width=.66\linewidth]{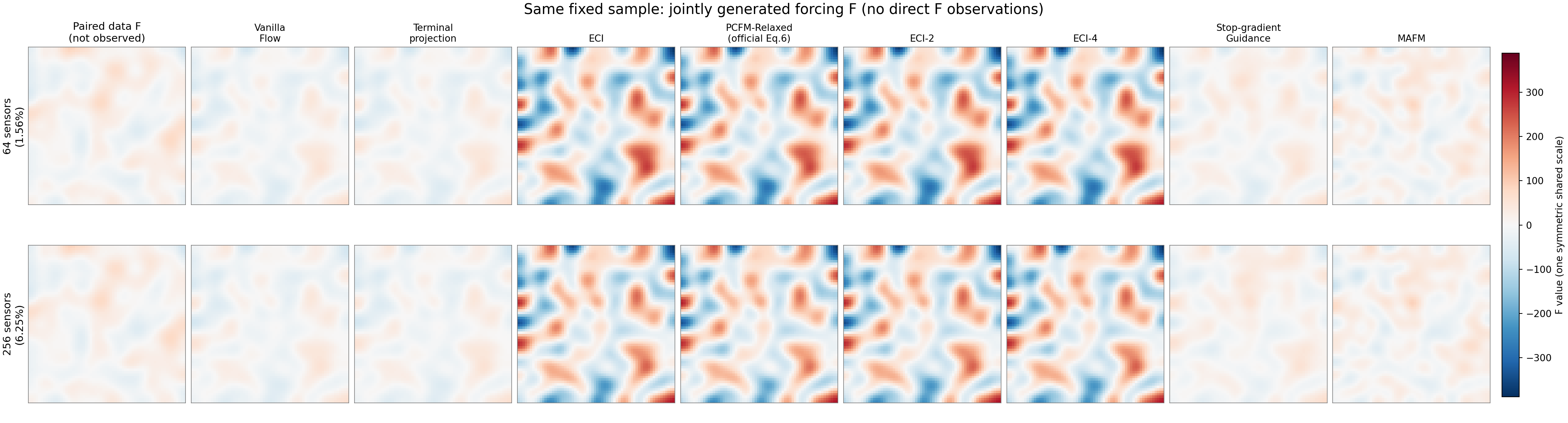}%
\hfill
\includegraphics[trim=1235.4bp 0 0 20bp,clip,width=.315\linewidth]{figures/poisson_samples_f.png}
\caption{Poisson source fields $F$ paired with Figure~\ref{fig:poisson-u-samples},
with the same representative ECI column.}
\label{fig:poisson-f-samples}
\end{figure}

\begin{figure}[!t]
\centering
\includegraphics[width=.96\linewidth]{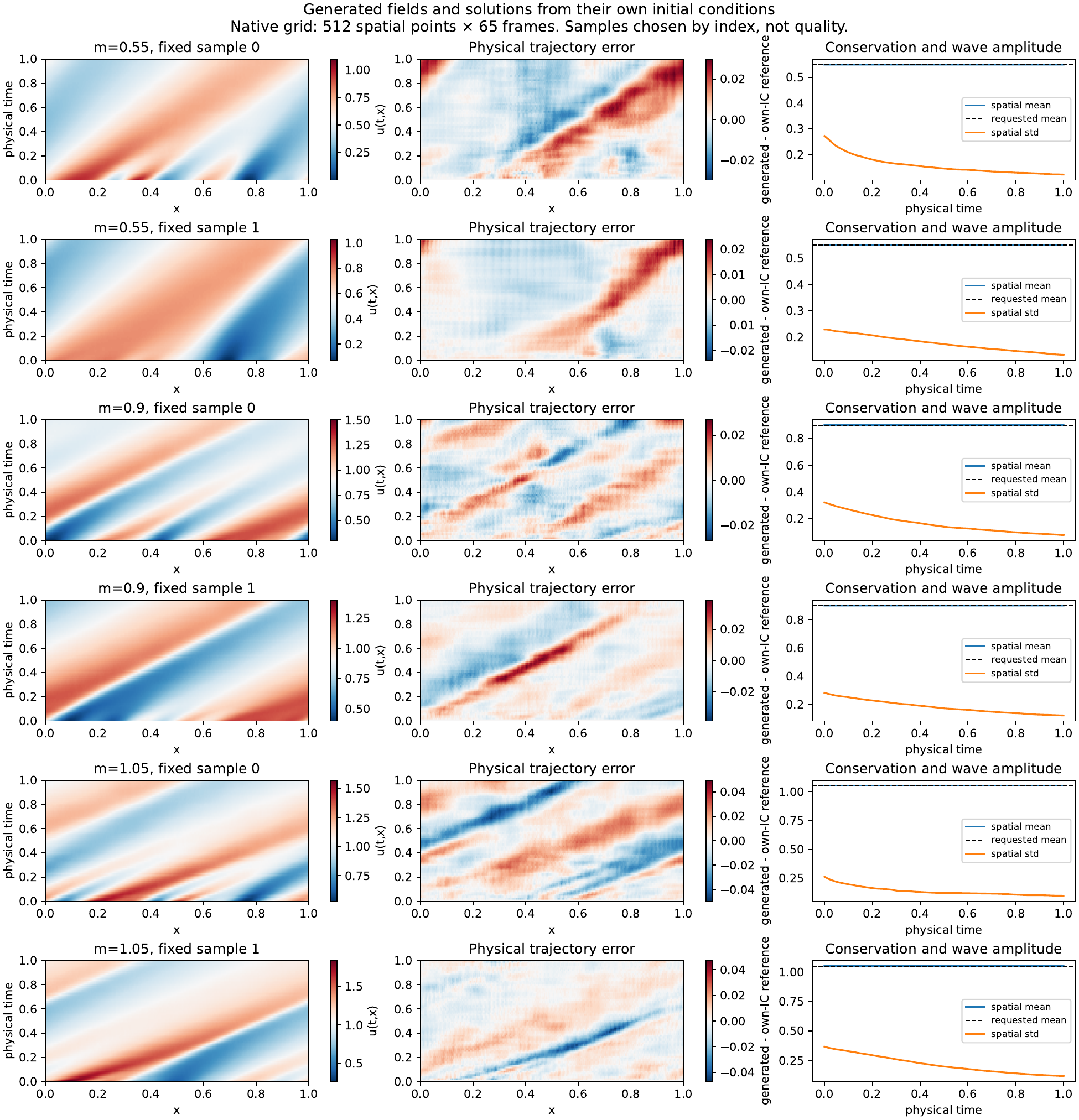}
\caption{Periodic Burgers samples.}
\label{fig:burgers-samples}
\end{figure}

\begin{figure}[!t]
\centering
\includegraphics[width=.96\linewidth]{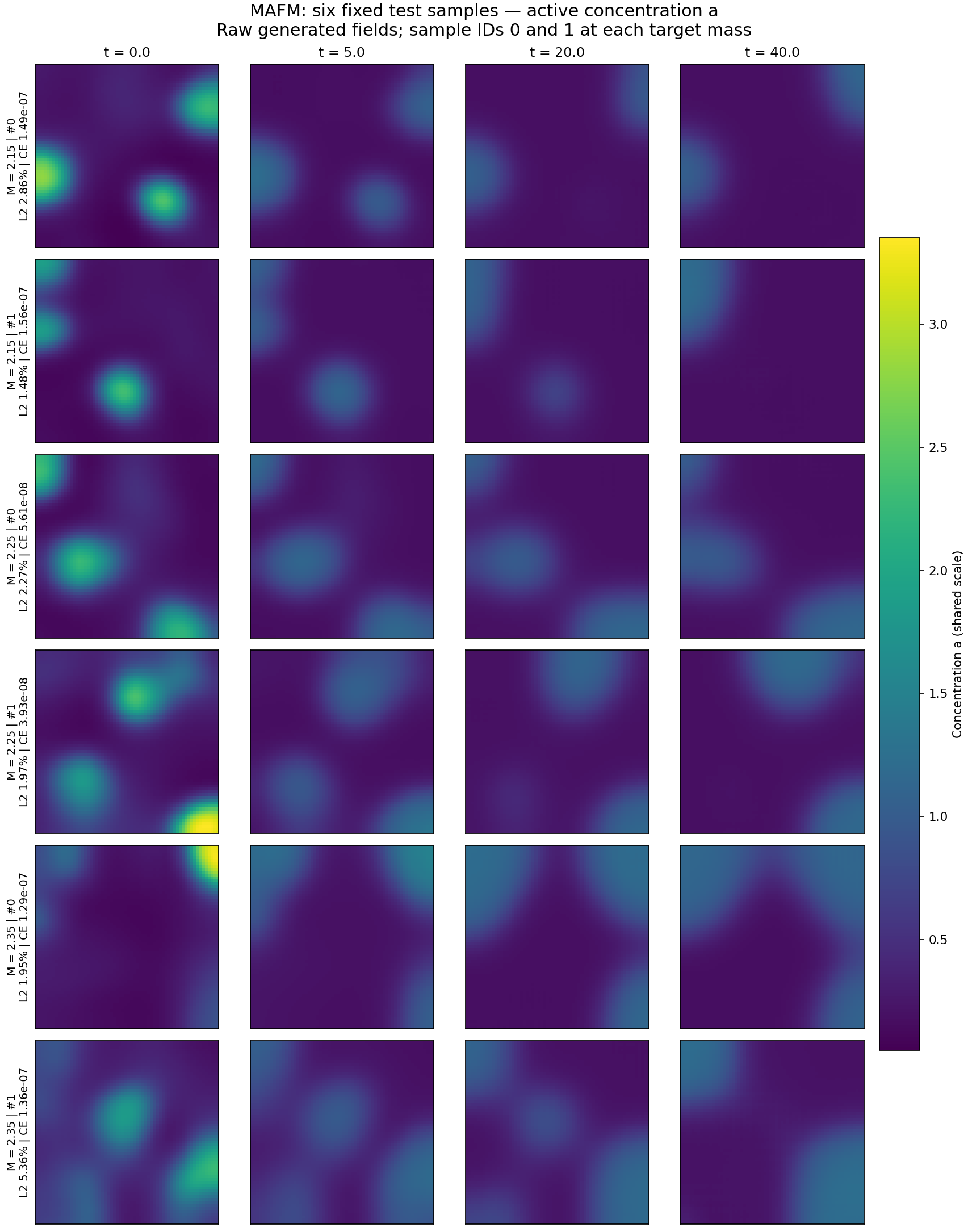}
\caption{Reaction--diffusion samples.}
\label{fig:reaction-diffusion-samples}
\end{figure}

\begin{figure}[!t]
\centering
\includegraphics[width=.96\linewidth]{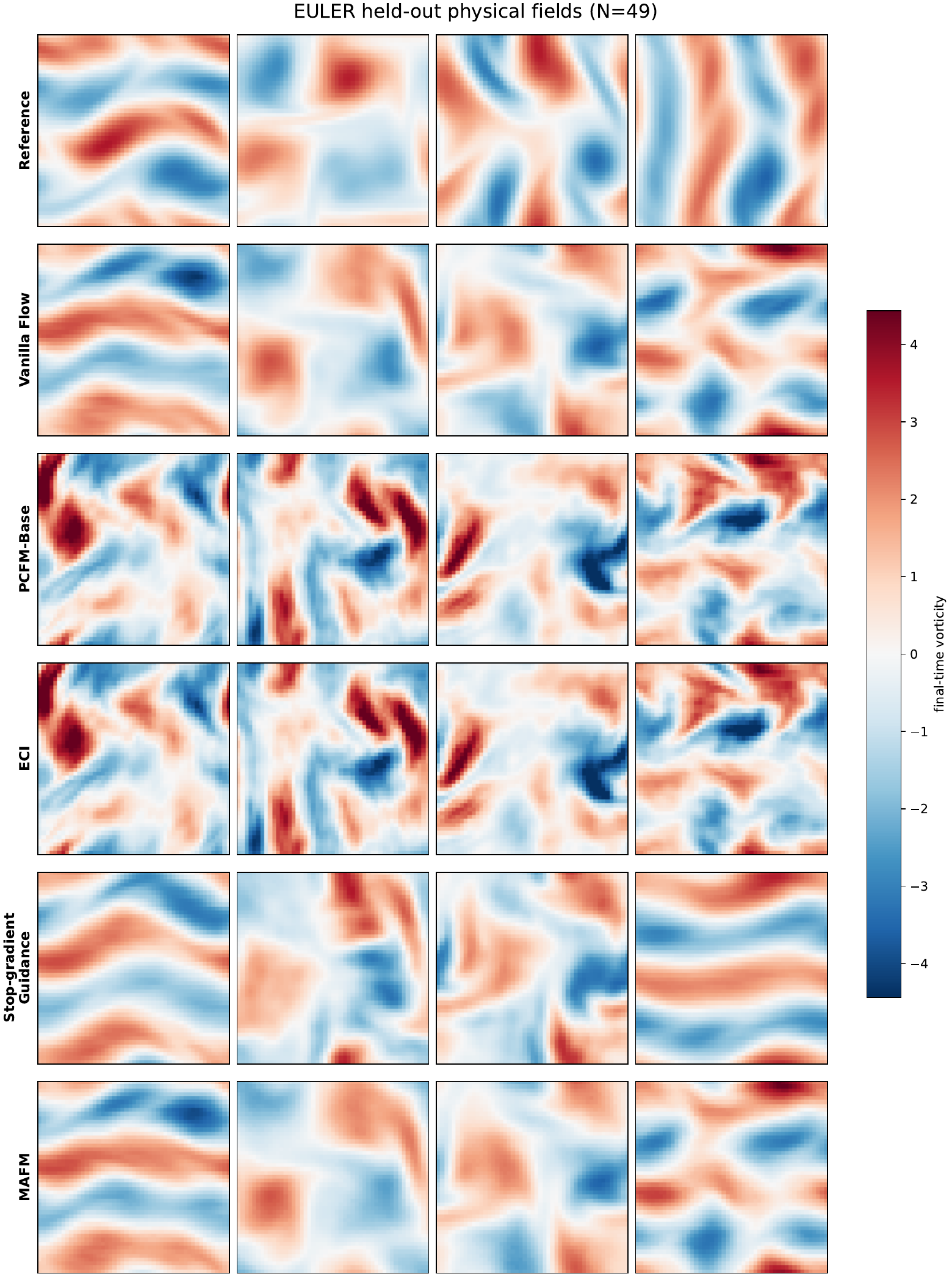}
\caption{Euler vorticity samples at fixed energy.}
\label{fig:euler-samples}
\end{figure}

\begin{figure}[!t]
\centering
\includegraphics[width=.90\linewidth]{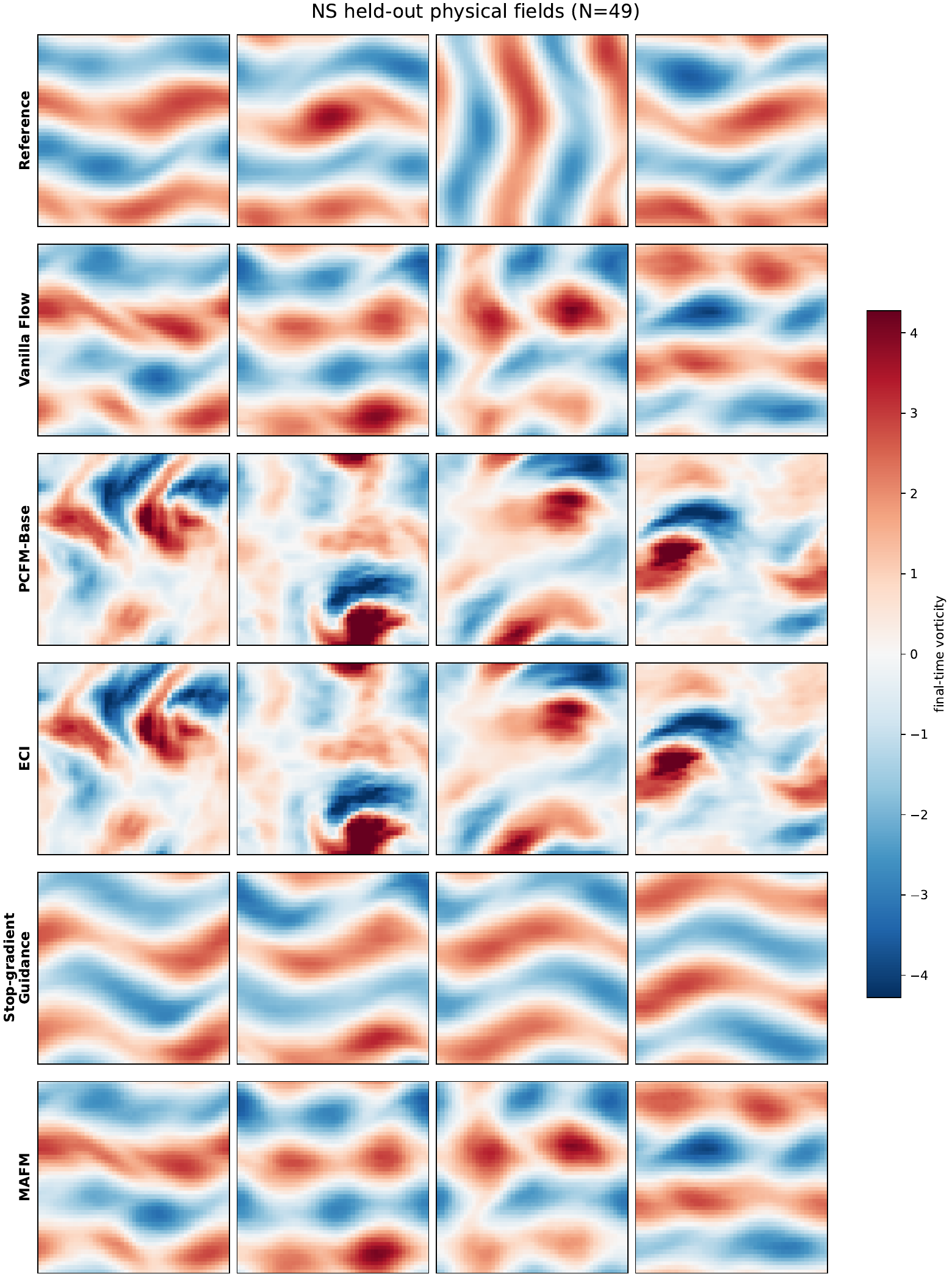}
\caption{Controlled-NS vorticity samples at fixed energy.}
\label{fig:ns-samples}
\end{figure}

\FloatBarrier
\subsection{\rev{Additional inference-time baselines}}
\label{app:additional-inverse-baselines}

\begin{revision}
Table~\ref{tab:additional-inverse-baselines} compares D-Flow
\citep{benhamu2024dflowdifferentiatingflowscontrolled}, PnP-Flow~\citep{martin2025pnpflowplugandplayimagerestoration}, and Lagrangian Dual Flows (LDF)~\citep{kurtz2026constrainedflowmatchinglagrangian} with CCFM and
MAFM on the same frozen checkpoints, paired test inputs, and prescribed
constraints. The two energy tests use 512 samples each. Burgers and
reaction--diffusion use 768 and 384 samples, respectively, with equal
weight for each of three target masses. Each Poisson sensor density uses
two layouts with 512 samples per layout.

The data-fidelity terms use the supplied mass or sensor equality, or the
target per-frame velocity radius for the energy equality. D-Flow optimizes
the source noise through the full 100-step generation ODE using L-BFGS
with a 20-iteration budget and a learning rate of $0.1$.
It uses 4{,}100 model forward evaluations (including recomputation) and
2{,}000 input VJPs per sample. PnP-Flow alternates a data-fidelity step,
interpolation with fresh prior noise, and a frozen endpoint prediction.
Its fidelity step size is $(1-t)^\alpha$, with $\alpha$ selected on a
separate development set. It uses 100 forward evaluations and no input VJPs.
We evaluate both methods' raw outputs.
LDF evolves the sample and equality multipliers with the explicit midpoint
method at the authors' default stepsize of $0.01$. We read the last saved
sample before $t=1$ after 198 forward evaluations, with terminal projection
disabled. We select the residual weight on independent development inputs
for each task. \clarrev{CCFM uses $p(s)=(s/2)^2$.} CCFM and both MAFM rows use 100 forward evaluations.
The projected MAFM row applies a terminal projection to the same samples
as the preceding row.

On Euler and controlled NS, MAFM gives the lowest relative $L_2$ and
divergence RMSE among the methods in this table. D-Flow or PnP-Flow give
lower relative $L_2$ on some other tasks with larger constraint errors.
Both miss the prescribed numerical feasibility thresholds in these
comparisons. MAFM gives lower constraint and PDE errors than LDF across
all six settings. Our full-gradient endpoint baseline tests the guidance
operation shared by FunDPS~\citep{yao2026guideddiffusionsamplingfunction} and DiffusionPDE
\citep{huang2024diffusionpde}, namely differentiating a loss on a denoised
prediction back to the current state. DiffusionPDE also evaluates an
explicit PDE residual in its guidance loss.
\end{revision}
\input{tables/additional_inverse_baselines}

%% file: tables/sphere_full.tex
\begin{table}[!htbp]
\caption{Complete sphere--plane results ($10{,}000$ samples).}
\label{tab:sphere-full}
\centering
\scriptsize
\setlength{\tabcolsep}{3.0pt}
\begin{tabular*}{\linewidth}{@{\extracolsep{\fill}}lcccccc@{}}
\toprule
Method & mean $|z|$ & q99 $|z|$ & Radial $W_1$ & SWD & MMD$^2$ & Joint \\
\midrule
Vanilla FM & 0.5009 & 0.2877 & 0.2158 & 0.1281 & 0.05516 & 0.06\% \\
Terminal projection & 0 & 0 & 0.2158 & 0.1078 & 0.03237 & 6.47\% \\
PCFM-Base & 0 & 0 & 0.006273 & 0.009357 & 0.000559 & 56.57\% \\
PCFM-Relaxed & $2.103\mathrm{e}{-5}$ & $1.358\mathrm{e}{-5}$ & 0.006328 & 0.009551 & 0.000562 & 56.08\% \\
ECI (operator-equivalent) & 0 & 0 & 0.006273 & 0.009357 & 0.000559 & 56.57\% \\
Stop-grad. guidance & 0 & 0 & 0.006310 & 0.01885 & 0.000768 & 57.35\% \\
Full-grad. guidance & $2.521\mathrm{e}{-6}$ & $2.122\mathrm{e}{-6}$ & 0.003888 & 0.02417 & 0.000930 & 70.52\% \\
MAFM & $3.372\mathrm{e}{-6}$ & $2.866\mathrm{e}{-6}$ & 0.004929 & 0.009568 & 0.000534 & 72.97\% \\
MAFM + projection & 0 & 0 & 0.004929 & 0.009568 & 0.000534 & 72.97\% \\
\bottomrule
\end{tabular*}
\end{table}

%% file: tables/poisson_full.tex
\begin{table}[!htbp]
\caption{Complete Poisson results at two sensor densities.}
\label{tab:poisson-full}
\centering
\scriptsize
\setlength{\tabcolsep}{3.8pt}
\begin{tabular*}{\linewidth}{@{\extracolsep{\fill}}lcccc@{}}
\toprule
& \multicolumn{2}{c}{64 sensors} & \multicolumn{2}{c}{256 sensors} \\
\cmidrule(lr){2-3}\cmidrule(l){4-5}
Method & CE & PDE RelL2 & CE & PDE RelL2 \\
\midrule
Vanilla FM & 1.395 & 1.128\% & 1.425 & 1.128\% \\
Terminal projection & 0 & 22.73\% & 0 & 46.31\% \\
PCFM-Base & 0 & 11.51\% & 0 & 18.05\% \\
PCFM-Relaxed & 0 & 10.33\% & 0 & 16.07\% \\
ECI & 0 & 11.51\% & 0 & 18.05\% \\
ECI-2 & 0 & 11.50\% & 0 & 18.14\% \\
ECI-4 & 0 & 11.50\% & 0 & 18.19\% \\
Stop-grad. guidance & $3.291\mathrm{e}{-9}$ & 5.554\% & $1.350\mathrm{e}{-9}$ & 5.636\% \\
Full-grad. guidance & 0.610 & 1.050\% & 0.371 & 0.974\% \\
MAFM & $1.721\mathrm{e}{-7}$ & 1.705\% & $1.751\mathrm{e}{-7}$ & 2.252\% \\
MAFM + projection & 0 & 1.705\% & 0 & 2.252\% \\
\bottomrule
\end{tabular*}
\end{table}

%% file: tables/conservation_full.tex
\begin{table}[H]
\caption{\textbf{Complete periodic Burgers results} ($768$ independent samples). CE is the mean-trajectory mass error. RelL2 and fluctuation RelL2 are percentages.
\rev{Relative std.\ error is the $L_2$ norm of the difference between the generated and reference pointwise standard-deviation fields, divided by the $L_2$ norm of the reference field.}}
\label{tab:burgers-full}
\centering
\scriptsize
\setlength{\tabcolsep}{3.2pt}
\begin{tabular*}{\linewidth}{@{\extracolsep{\fill}}lccccc@{}}
\toprule
Method & RelL2 & CE & Fluct. RelL2 & Field SWD & \rev{Relative std.\ error} \\
\midrule
Vanilla FM & 1.134\% & $3.565\mathrm{e}{-1}$ & 3.163\% & 0.2695 & 0.2323 \\
Terminal projection & 26.97\% & $5.490\mathrm{e}{-8}$ & 74.48\% & 0.07782 & 0.0883 \\
PCFM-Base & 7.610\% & $5.392\mathrm{e}{-8}$ & 13.48\% & 0.1776 & 0.5160 \\
PCFM-Relaxed & 7.610\% & $3.169\mathrm{e}{-6}$ & 13.47\% & 0.1777 & 0.5163 \\
ECI (operator-equivalent) & 7.610\% & $5.392\mathrm{e}{-8}$ & 13.48\% & 0.1776 & 0.5160 \\
ECI-2 & 7.537\% & $5.397\mathrm{e}{-8}$ & 13.38\% & 0.1766 & 0.5125 \\
ECI-4 & 7.890\% & $5.396\mathrm{e}{-8}$ & 13.93\% & 0.1834 & 0.5333 \\
ECI-5 (resampled) & 2.708\% & $5.370\mathrm{e}{-8}$ & 6.965\% & 0.06150 & 0.1347 \\
Stop-grad. guidance & 1.030\% & $5.382\mathrm{e}{-8}$ & 16.80\% & 0.2876 & 0.8484 \\
Full-grad. guidance & 17.18\% & $2.634\mathrm{e}{-2}$ & 23.78\% & 0.2901 & 0.8569 \\
MAFM & 0.9648\% & $4.216\mathrm{e}{-8}$ & 4.828\% & 0.1574 & 0.4556 \\
MAFM + projection & 0.9648\% & $5.320\mathrm{e}{-8}$ & 4.828\% & 0.1574 & 0.4556 \\
\bottomrule
\end{tabular*}
\end{table}

\begin{table}[H]
\caption{\textbf{Complete mass-conserving reaction--diffusion results} ($384$ independent samples). RelL2 and q95 RelL2 are percentages.}
\label{tab:rd-full}
\centering
\scriptsize
\setlength{\tabcolsep}{3.0pt}
\begin{tabular*}{\linewidth}{@{\extracolsep{\fill}}lccccc@{}}
\toprule
Method & Field SWD & RelL2 & CE & q95 RelL2 & Negative fraction \\
\midrule
Vanilla FM & 0.09650 & 1.312\% & $5.501\mathrm{e}{-2}$ & 5.201\% & $4.48\mathrm{e}{-6}$ \\
Terminal projection & 0.08529 & 10.39\% & $5.837\mathrm{e}{-8}$ & 41.51\% & $8.75\mathrm{e}{-5}$ \\
PCFM-Base & 0.1303 & 3.030\% & $5.894\mathrm{e}{-8}$ & 10.26\% & $1.79\mathrm{e}{-3}$ \\
PCFM-Relaxed & 0.1157 & 2.458\% & $3.720\mathrm{e}{-4}$ & 8.482\% & $1.37\mathrm{e}{-3}$ \\
ECI & 0.1246 & 2.962\% & $5.855\mathrm{e}{-8}$ & 10.18\% & $1.78\mathrm{e}{-3}$ \\
Stop-grad. guidance & 0.1220 & 6.040\% & $5.922\mathrm{e}{-8}$ & 23.45\% & $1.53\mathrm{e}{-6}$ \\
Full-grad. guidance & 0.06375 & 1.563\% & $1.695\mathrm{e}{-3}$ & 5.878\% & $4.88\mathrm{e}{-6}$ \\
MAFM & 0.06659 & 2.740\% & $1.134\mathrm{e}{-7}$ & 11.07\% & $2.33\mathrm{e}{-4}$ \\
MAFM + projection & 0.06659 & 2.740\% & $6.773\mathrm{e}{-8}$ & 11.07\% & $2.33\mathrm{e}{-4}$ \\
\bottomrule
\end{tabular*}
\end{table}

%% file: tables/core_ablation_full.tex
\begin{table}[!htbp]
\centering
\scriptsize
\setlength{\tabcolsep}{4.5pt}
\renewcommand{\arraystretch}{1.04}
\resizebox{\textwidth}{!}{%
\begin{tabular}{llrrrrl}
\toprule
Task & Variant & CE & Physical RelL2 & Field SWD & $\Delta$ RelL2 & Paired 95\% CI \\
\midrule
Poisson & Full MAFM
  & $1.721\mathrm{e}{-7}$ & 0.01705 & -- & -- & -- \\
 & w/o adjoint
  & $1.721\mathrm{e}{-7}$ & 0.3803 & -- & 0.3632 & [0.3521, 0.3751] \\
 & \costrev{w/o support}
  & \costrev{$1.721\mathrm{e}{-7}$} & \costrev{0.2899} & \costrev{--}
  & \costrev{0.2728} & \costrev{[0.2534, 0.2937]} \\
\midrule
Burgers & Full MAFM
  & $4.216\mathrm{e}{-8}$ & 0.009648 & 0.1574 & -- & -- \\
 & Reset anchor
  & 0.008997 & 0.09437 & 0.1767 & 0.08473 & [0.07948, 0.09028] \\
 & w/o support
  & $1.181\mathrm{e}{-7}$ & 0.02703 & 0.3135 & 0.01738 & [0.01652, 0.01828] \\
 & w/o normal
  & 0.2698 & 0.01286 & 0.2750 & 0.003207 & [0.002873, 0.003559] \\
\midrule
Reaction--diffusion & Full MAFM
  & $1.134\mathrm{e}{-7}$ & 0.02740 & 0.06659 & -- & -- \\
 & Reset anchor
  & 0.04413 & 0.06145 & 0.2110 & 0.03405 & [0.02654, 0.04189] \\
 & w/o support
  & $1.147\mathrm{e}{-7}$ & 0.1058 & 0.2102 & 0.07839 & [0.07624, 0.08044] \\
 & w/o normal
  & 0.03186 & 0.02190 & 0.09998 & $-0.005500$ & [$-0.007283$, $-0.003750$] \\
\bottomrule
\end{tabular}
}
\caption{Core-ablation aggregates with paired confidence intervals.}
\label{tab:core-ablation-full}
\end{table}

%% file: tables/oracle_full.tex
\begin{table}[!htbp]
\caption{Three-constraint oracle comparison for Euler and controlled NS (64 samples each).}
\label{tab:oracle-full}
\centering
\scriptsize
\setlength{\tabcolsep}{2.5pt}
\begin{tabular*}{\linewidth}{@{\extracolsep{\fill}}llcccccc@{}}
\toprule
Case & Method & Feature SWD & Field SWD & RelL2 & Div. & Mean & Energy \\
\midrule
Euler & Vanilla FM & 0.152 & 0.0662 & 0.0368 & 0.0635 & 0.0131 & 0.0186 \\
& Terminal projection & 0.152 & 0.0663 & 0.0332 & $1.45\times 10^{-6}$ & $2.86\times 10^{-8}$ & $4.78\times 10^{-7}$ \\
& PCFM-Base & 0.424 & 0.115 & 0.125 & $1.48\times 10^{-6}$ & $1.65\times 10^{-8}$ & $4.77\times 10^{-7}$ \\
& PCFM-Relaxed & 0.427 & 0.116 & 0.124 & 0.00766 & $9.65\times 10^{-4}$ & 0.00182 \\
& ECI (operator-equivalent) & 0.424 & 0.115 & 0.125 & $1.48\times 10^{-6}$ & $1.65\times 10^{-8}$ & $4.77\times 10^{-7}$ \\
& Stop-grad. guidance & 0.427 & 0.0985 & 0.0429 & 0.0598 & 0.0161 & 0.0262 \\
& Full-grad. guidance & 0.184 & 0.0667 & 0.0341 & 0.0602 & 0.0106 & 0.0160 \\
& MAFM & 0.157 & 0.0676 & 0.0320 & $1.62\times 10^{-6}$ & $1.86\times 10^{-7}$ & $7.15\times 10^{-6}$ \\
& MAFM + projection & 0.157 & 0.0676 & 0.0320 & $1.44\times 10^{-6}$ & $2.22\times 10^{-8}$ & $4.78\times 10^{-7}$ \\
\midrule
NS & Vanilla FM & 0.0690 & 0.0495 & 0.0346 & 0.0624 & 0.0111 & 0.0181 \\
& Terminal projection & 0.0634 & 0.0495 & 0.0309 & $1.45\times 10^{-6}$ & $2.22\times 10^{-8}$ & $4.77\times 10^{-7}$ \\
& PCFM-Base & 0.375 & 0.0943 & 0.112 & $1.47\times 10^{-6}$ & $1.79\times 10^{-8}$ & $4.73\times 10^{-7}$ \\
& PCFM-Relaxed & 0.387 & 0.0947 & 0.112 & 0.00707 & $9.61\times 10^{-4}$ & 0.00179 \\
& ECI (operator-equivalent) & 0.375 & 0.0943 & 0.112 & $1.47\times 10^{-6}$ & $1.79\times 10^{-8}$ & $4.73\times 10^{-7}$ \\
& Stop-grad. guidance & 0.405 & 0.0797 & 0.0355 & 0.0486 & 0.00878 & 0.0201 \\
& Full-grad. guidance & 0.252 & 0.0622 & 0.0289 & 0.0504 & 0.00692 & 0.0173 \\
& MAFM & 0.0961 & 0.0506 & 0.0294 & $1.62\times 10^{-6}$ & $1.89\times 10^{-7}$ & $7.18\times 10^{-6}$ \\
& MAFM + projection & 0.0961 & 0.0506 & 0.0294 & $1.44\times 10^{-6}$ & $2.14\times 10^{-8}$ & $4.66\times 10^{-7}$ \\
\bottomrule
\end{tabular*}
\end{table}

%% file: sections/rebuttal_results.tex
\subsection{\rev{Field SWD and matched CCFM comparisons}}
\label{app:field-swd}
\label{app:ccfm-burgers}
\begin{revision}
\paragraph{Field SWD.}
We flatten each full trajectory in the generated and reference ensembles
and project both onto the same fixed directions. For direction $j$, let
$g_{(i)j}$ and $r_{(i)j}$ be the sorted projections of the generated and
reference fields. For equal ensemble sizes $n$ and $L$ directions, the
reported distance is
\[
  \mathrm{Field\ SWD}=
  \left[\frac{1}{nL}\sum_{j=1}^L\sum_{i=1}^n
  \bigl(g_{(i)j}-r_{(i)j}\bigr)^2\right]^{1/2}.
\]
This diagnostic measures agreement with the chosen reference ensemble along
the evaluated directions. It is defined even when the reference law is
singular. The constraint specifies the feasible set and the reference
sampling protocol specifies how to sample on it. For Burgers and
reaction--diffusion (Table~\ref{tab:swd-conservation}), the reference
trajectories are generated independently at each prescribed mass
(Appendices~\ref{app:burgers-protocol} and~\ref{app:rd-protocol}).
For fixed-energy Euler and controlled NS (Table~\ref{tab:swd-energy}),
we use held-out trajectories at the training energy.
Own-IC relative $L_2$ uses a separate solver reference for each generated
trajectory.

Tables~\ref{tab:swd-conservation} and~\ref{tab:swd-energy} retain the Field SWD values from the conservation and energy-only comparisons. On Burgers, resampled ECI has mean Field SWD 0.06150 and own-IC 2.708\%, compared with 0.1574 and 0.965\% for MAFM. The sphere--plane diagnostics in Appendix~\ref{app:complete-results} retain the distinct radial and full-field comparisons.

\clarrev{We adapt CCFM's Gaussian projection to each frame's mass equality by imposing two one-sided marginal chance constraints. At candidate time $s=(k+1)/100$, the two schedules use $p=(s/2)^n$ for $n=1,2$. With $512$ independent standard Gaussian source entries per frame, we clamp the candidate frame mean to a band centered at $s m_\star$ with half-width $-(1-s)\Phi^{-1}(p)/\sqrt{512}$, where $\Phi$ is the standard normal CDF. The band reduces to exact mass at $s=1$. Each method uses the same frozen model and 768 paired sources (256 per mass) at 100 forward evaluations. This is a Burgers mass-equality adaptation of CCFM. Its published NS setup supplies an initial condition and a global conservation constraint.}
\end{revision}

\begin{table}[H]
\centering\small
\caption{\clarrev{Burgers CCFM comparison by target mass. Own-IC is relative $L_2$ in percent. The last three rows average the three prescribed masses.}}
\label{tab:ccfm-burgers}
{\color{clarityink}
\begin{tabular}{llccc}
\toprule
Mass & Sampler & CE & Own-IC (\%) & Field SWD \\
\midrule
0.55 & CCFM ($n=1$) & $5.316\times10^{-8}$ & 5.394 & 0.06016 \\
     & CCFM ($n=2$) & $5.357\times10^{-8}$ & 7.563 & 0.06381 \\
     & MAFM & $2.661\times10^{-8}$ & 1.092 & 0.16484 \\
0.90 & CCFM ($n=1$) & $4.043\times10^{-8}$ & 3.365 & 0.05831 \\
     & CCFM ($n=2$) & $4.020\times10^{-8}$ & 4.848 & 0.05910 \\
     & MAFM & $4.853\times10^{-8}$ & 0.829 & 0.14856 \\
1.05 & CCFM ($n=1$) & $6.809\times10^{-8}$ & 3.491 & 0.07068 \\
     & CCFM ($n=2$) & $6.803\times10^{-8}$ & 4.776 & 0.07377 \\
     & MAFM & $5.133\times10^{-8}$ & 0.973 & 0.15871 \\
\midrule
Mean & CCFM ($n=1$) & $5.39\times10^{-8}$ & 4.083 & 0.06305 \\
     & CCFM ($n=2$) & $5.39\times10^{-8}$ & 5.729 & 0.06556 \\
     & MAFM & $4.22\times10^{-8}$ & 0.965 & 0.15737 \\
\bottomrule
\end{tabular}
}
\end{table}
\begin{revision}[clarityink]
Table~\ref{tab:ccfm-five} extends the comparison to the other PDE benchmarks using their original frozen models, paired sources and 100 forward calls. Reaction--diffusion uses the affine total-mass condition, and Poisson uses affine sensor selection. For Euler and controlled NS, we linearize the framewise energy equality at the predicted endpoint and use the variance along that affine constraint under the original Gaussian source. Each CCFM run applies an exact terminal projection for its supplied constraint. Table~\ref{tab:dynamics-main} reports the common $n=2$ schedule across tasks.
\end{revision}
\input{tables/ccfm_comparison}
\input{tables/field_swd_archive}

\subsection{\rev{Burgers tangential history control}}
\label{app:anchor-control}
\begin{revision}
The two arms share the frozen model, 768 paired sources, initial anchor, normal contraction and 100 forward calls. The intervention sets $\rho_k=1$ for the tangential update while retaining the affine normal update. The maximum difference between paired per-step mass residuals is $3.58\times10^{-7}$. CE is $4.22\times10^{-8}$ with finite tracking and $4.86\times10^{-8}$ with full replacement. Own-IC is 0.965\% and 7.68\%, respectively, with a paired difference of 6.72 percentage points (95\% interval $[6.36,7.08]$, from 10,000 paired bootstrap draws stratified by mass). Field SWD is 0.1574 and 0.1776. The original reset in Table~\ref{tab:core-ablation} also changes feasibility.
Setting $\rho_k=1$ in Eq.~\eqref{eq:anchor-memory} sets the weight of the
previous tangential anchor to zero at that step. In the reported affine
implementation, both arms use the same exponential normal substep, which
contracts the same initial mass residual by the same factor at each step.
This accounts for the comparable mass enforcement in
Section~\ref{sec:core-ablation}.
\end{revision}

\subsection{\rev{PCFM projection and mixing controls}}
\label{app:pcfm-mixing-controls}
\begin{revision}
Each case uses the same 512 paired sources as Table~\ref{tab:dynamics-main}c and the same frozen model. PCFM-Relaxed uses Eq.~6-style refinement with 20 inner steps and penalty weight 1. The added terminal projection acts on the exact replayed trajectories. The two-mix controls use 50 steps and two forward calls per step. In the noise-refresh arm, the reinterpolation source is redrawn from the original prior every fifth call. Both mixing arms use the same energy-only terminal projection. The 95\% paired intervals for refreshed minus fixed own-IC are $[-0.25,0.48]$ percentage points on Euler and $[-0.20,0.38]$ on controlled NS. The official PCFM operator equivalence check is limited to the Poisson sparse-selector setting, where the stepwise and whole-sampler tests gave zero maximum absolute difference.

The ECI row in Table~\ref{tab:dynamics-main}(c) reuses the one-mix
update equivalent to PCFM-Base in this setting. Terminal projection of
vanilla starts from trajectories whose own-IC errors are already close to
their post-projection values. PCFM corrects intermediate predictions, which
also affect later model queries. The observed rollout gap persists under
both mixing schedules and with comparable final energy errors. These
controls do not separately identify the effects of per-step correction
strength and projection schedule.

The constraint error in Appendix~J of PCFM~\citep{utkarsh2026physics}
is the $\ell_2$ norm of a prescribed residual, averaged over samples.
The PCFM ensemble metrics compare generated and reference sets. Here
own-IC relative $L_2$ compares each generated trajectory with an independent
PDE rollout initialized from its own generated first frame, prepared as in
Appendices~\ref{app:euler-protocol} and~\ref{app:ns-protocol}. A trajectory
can therefore have small energy CE and a large rollout error.

The initial-frame RMS correction from the generated field to the solver input is 0.03755 and 0.03509 for PCFM-Base on Euler and controlled NS, versus 0.01103 and 0.01076 for terminal projection of vanilla. These are diagnostics of the generated first frame, and the own-IC comparison uses the unmodified future frames.
\end{revision}

\begin{table}[H]
\centering\small
\caption{\rev{Fixed-energy control runs. Own-IC is relative $L_2$ in percent.}}
\label{tab:pcfm-mixing-controls}
{\color{revisionink}
\begin{tabular}{llrrrr}
\toprule
Case & Sampler & Energy CE & Own-IC (\%) & Divergence & Field SWD \\
\midrule
Euler & PCFM-Relaxed replay & $1.84\times10^{-3}$ & 13.56 & 0.2285 & 0.11503 \\
& + terminal projection & $2.34\times10^{-7}$ & 13.56 & 0.2287 & 0.11506 \\
& Two mix, fixed source & $1.47\times10^{-7}$ & 13.41 & 0.2266 & 0.11407 \\
& Two mix, refreshed source & $1.50\times10^{-7}$ & 13.51 & 0.2303 & 0.09906 \\
\midrule
Controlled NS & PCFM-Relaxed replay & $1.81\times10^{-3}$ & 12.29 & 0.2123 & 0.09479 \\
& + terminal projection & $2.44\times10^{-7}$ & 12.29 & 0.2124 & 0.09485 \\
& Two mix, fixed source & $1.52\times10^{-7}$ & 12.13 & 0.2086 & 0.09376 \\
& Two mix, refreshed source & $1.47\times10^{-7}$ & 12.21 & 0.2168 & 0.09892 \\
\bottomrule
\end{tabular}
}
\end{table}

\subsection{\rev{Shifted energy targets}}
\label{app:shifted-energy}
\begin{revision}
We use 128 paired sources per case and target, the original frozen checkpoints, and the fixed training-energy baseline settings. We adjust independent held-out initial fields to each target and evolve them with the original solver for ensemble evaluation. Own-IC uses a separate solver rollout from each generated sample's initial field prepared at the target energy. For PCFM-Relaxed and MAFM, Table~\ref{tab:shifted-energy} gives the same generated outputs before and after a shared framewise radial projection. On the shifted PCFM outputs, the radial solution agrees with the official Newton--Schur terminal solver to within $9\times10^{-16}$ in float64. The table evaluates the common float32 radial projection for both methods.
\end{revision}
\input{tables/shifted_energy_full}

\begin{revision}
For a compact comparison with one setting per baseline, we select the
ECI and CCFM settings by their equally weighted mean relative $L_2$ errors
across the four $0.8\times$ and $1.2\times$ system--target pairs in
Table~\ref{tab:shifted-energy}.
For ECI, two mixes with source refresh give $13.23\%$, compared with
$13.49\%$ for two mixes with fixed source and $13.74\%$ for one mix.
For CCFM, $n=2$ gives $4.27\%$ and $n=1$ gives $4.89\%$.
Across the four settings, MAFM's mean energy error ranges from
$1.26\times10^{-5}$ to $2.07\times10^{-5}$ before projection and from
$2.10\times10^{-7}$ to $2.24\times10^{-7}$ after the shared projection.
The projected row has the lowest relative $L_2$ error among feasible
methods in all four settings, as reported in Section~\ref{sec:energy}.
\end{revision}

\begin{revision}
\paragraph{Doubled energy.}
Table~\ref{tab:doubled-energy} compares five methods at twice the
training energy with 128 paired sources per system and 100 forward calls
per sample. We retain the ECI and CCFM configurations selected at the two
moderate shifts. With the shared terminal projection, MAFM meets the energy
tolerance and has the lowest relative $L_2$ among the feasible methods shown.
Its rollout errors increase to $9.85\%$ on Euler and $10.01\%$ on controlled
NS. Doubling kinetic energy requires increasing the velocity norm by about
$41\%$. The result is consistent with less accurate endpoint guidance under
a larger shift, even though terminal projection still enforces the energy
target.
\end{revision}
\input{tables/doubled_energy}

\subsection{\costrev{Normal-contraction schedules}}
\label{app:normal-schedules}
\begin{revision}[costink]
In the ideal dynamics, $\int_0^1\kappa(t)\,\mathrm{d}t$ sets the cumulative log contraction of the constraint residual. The discrete normal budgets for sphere--plane, Poisson, Burgers and reaction--diffusion are $\log(10^4)$, $\log(10^8)$, $32$ and $\log(10^6)$, respectively. For Euler and controlled NS the budget is $\log(10^5)$, applied to the framewise radius deviation during the radial substep after tangential tracking. With a constant budget $K$ over $N$ equal steps, the exponential normal substep multiplies the relevant residual or radius deviation by $\exp(-K/N)$ at each step. For Burgers, $K=32$ over 100 steps gives
a per-step factor of $\exp(-32/100)\simeq0.726$. The rate $\kappa(t)$
sets how quickly feasibility is approached. Rapid attraction relative to
tangential motion can require smaller steps in explicit integration.
\end{revision}

%% file: tables/ccfm_comparison.tex
\begin{table}[H]
\centering\small
\caption{\clarrev{CCFM schedule comparison across the PDE benchmarks. The relative $L_2$ errors use own-initial-condition rollouts for time-dependent PDEs and an independent solve from the generated source for Poisson.}}
\label{tab:ccfm-five}
{\color{clarityink}
\begin{tabular}{lrrrr}
\toprule
 & \multicolumn{2}{c}{$n=1$} & \multicolumn{2}{c}{$n=2$} \\
\cmidrule(lr){2-3}\cmidrule(lr){4-5}
Benchmark & CE & Relative $L_2$ & CE & Relative $L_2$ \\
\midrule
Controlled NS & \ensuremath{2.35\mathrm{e}{-7}} & 4.05\% & \ensuremath{2.35\mathrm{e}{-7}} & 3.48\% \\
Euler & \ensuremath{2.39\mathrm{e}{-7}} & 4.20\% & \ensuremath{2.33\mathrm{e}{-7}} & 3.68\% \\
Burgers & \ensuremath{5.39\mathrm{e}{-8}} & 4.08\% & \ensuremath{5.39\mathrm{e}{-8}} & 5.73\% \\
Reaction--diffusion & \ensuremath{5.92\mathrm{e}{-8}} & 6.49\% & \ensuremath{5.96\mathrm{e}{-8}} & 6.83\% \\
Poisson (64 sensors) & 0 & 22.74\% & 0 & 22.78\% \\
Poisson (256 sensors) & 0 & 45.45\% & 0 & 46.47\% \\
\bottomrule
\end{tabular}
}
\end{table}

%% file: tables/field_swd_archive.tex
\begin{table}[H]
\centering\small
\caption{\rev{Field SWD for the conservation rows in Table~\ref{tab:dynamics-main}b.}}
\label{tab:swd-conservation}
{\color{revisionink}
\begin{tabular}{lcc}
\toprule
Method & Burgers & Reaction--diffusion \\
\midrule
Vanilla FM & 0.270 & 0.0965 \\
FM + projection & 0.0778 & 0.0853 \\
PCFM-Base & 0.178 & 0.130 \\
PCFM-Relaxed & 0.178 & 0.116 \\
ECI$^\ast$ & 0.177 & 0.125 \\
Stop-grad. & 0.288 & 0.122 \\
Full-grad. (DPDE) & 0.290 & 0.0637 \\
MAFM & 0.157 & 0.0666 \\
MAFM + projection & 0.157 & 0.0666 \\
\bottomrule
\end{tabular}
}
\end{table}
\begin{table}[H]
\centering\small
\caption{\rev{Field SWD for the original submission's fixed-energy comparisons.}}
\label{tab:swd-energy}
{\color{revisionink}
\begin{tabular}{lcc}
\toprule
Method & Euler & Controlled NS \\
\midrule
Vanilla FM & 0.0662 & 0.0495 \\
Terminal projection & 0.0662 & 0.0495 \\
PCFM-Base & 0.114 & 0.0945 \\
PCFM-Relaxed & 0.115 & 0.0948 \\
ECI$^\dagger$ & 0.114 & 0.0945 \\
Stop-grad. guidance & 0.118 & 0.0792 \\
Full-grad. (DPDE) & 0.0667 & 0.0630 \\
MAFM & 0.0712 & 0.0533 \\
MAFM + projection & 0.0712 & 0.0533 \\
\bottomrule
\end{tabular}
}
\end{table}

%% file: tables/shifted_energy_full.tex
\begin{table}[H]
\centering\scriptsize
\setlength{\tabcolsep}{2pt}
\caption{\rev{Full comparisons at shifted energy targets (128 paired samples per target). Energy feasibility uses mean CE $\leq10^{-5}$. The two stop-gradient rows report their actual forward counts. All other rows use 100 forwards, including those that reuse a parent sampler and add a terminal projection.}}
\label{tab:shifted-energy}
{\color{revisionink}
\begin{tabular*}{\linewidth}{@{\extracolsep{\fill}}lrrrrrr@{}}
\toprule
\multicolumn{7}{l}{\textbf{Euler}} \\
& \multicolumn{3}{c}{$0.8\times$ training energy} & \multicolumn{3}{c}{$1.2\times$ training energy} \\
\cmidrule(lr){2-4}\cmidrule(lr){5-7}
Method & Energy CE & IC RelL2 & Div. RMS & Energy CE & IC RelL2 & Div. RMS \\
\midrule
Vanilla FM & \ensuremath{2.56\mathrm{e}{-1}} & 12.49\% & 0.0621 & \ensuremath{1.75\mathrm{e}{-1}} & 9.93\% & 0.0621 \\
Vanilla + projection & \ensuremath{2.07\mathrm{e}{-7}} & 4.01\% & 0.0556 & \ensuremath{2.32\mathrm{e}{-7}} & 4.60\% & 0.0682 \\
PCFM-Base / ECI-1 & \ensuremath{2.10\mathrm{e}{-7}} & 11.44\% & 0.1810 & \ensuremath{2.21\mathrm{e}{-7}} & 19.79\% & 0.3544 \\
PCFM-Relaxed (pre-proj.) & \ensuremath{1.94\mathrm{e}{-3}} & 11.75\% & 0.1884 & \ensuremath{3.56\mathrm{e}{-3}} & 20.44\% & 0.3642 \\
\quad + terminal projection & \ensuremath{2.23\mathrm{e}{-7}} & 11.76\% & 0.1885 & \ensuremath{2.24\mathrm{e}{-7}} & 20.44\% & 0.3642 \\
ECI-2 (fixed source) & \ensuremath{2.16\mathrm{e}{-7}} & 11.33\% & 0.1777 & \ensuremath{1.97\mathrm{e}{-7}} & 19.36\% & 0.3460 \\
ECI-2 (refreshed source) & \ensuremath{2.09\mathrm{e}{-7}} & 11.84\% & 0.1933 & \ensuremath{2.01\mathrm{e}{-7}} & 18.24\% & 0.3333 \\
Stop-grad. (199 calls) & \ensuremath{3.14\mathrm{e}{-2}} & 5.82\% & 0.0677 & \ensuremath{1.28\mathrm{e}{-1}} & 13.36\% & 0.2080 \\
Stop-grad. (99 calls) & \ensuremath{3.19\mathrm{e}{-2}} & 5.33\% & 0.0603 & \ensuremath{1.30\mathrm{e}{-1}} & 12.54\% & 0.1904 \\
Full-gradient guidance & \ensuremath{2.45\mathrm{e}{-1}} & 12.01\% & 0.0659 & \ensuremath{1.69\mathrm{e}{-1}} & 9.57\% & 0.0596 \\
CCFM ($n=1$) & \ensuremath{2.05\mathrm{e}{-7}} & 4.70\% & 0.0646 & \ensuremath{2.38\mathrm{e}{-7}} & 5.41\% & 0.1102 \\
CCFM ($n=2$) & \ensuremath{2.10\mathrm{e}{-7}} & 4.01\% & 0.0565 & \ensuremath{2.27\mathrm{e}{-7}} & 4.82\% & 0.1027 \\
MAFM (pre-proj.) & \ensuremath{1.93\mathrm{e}{-5}} & 2.96\% & 0.0400 & \ensuremath{1.26\mathrm{e}{-5}} & 3.60\% & 0.0666 \\
\quad + terminal projection & \ensuremath{2.11\mathrm{e}{-7}} & \textbf{2.96\%} & 0.0400 & \ensuremath{2.24\mathrm{e}{-7}} & \textbf{3.60\%} & 0.0666 \\
\bottomrule
\end{tabular*}
\par\smallskip
\begin{tabular*}{\linewidth}{@{\extracolsep{\fill}}lrrrrrr@{}}
\toprule
\multicolumn{7}{l}{\textbf{Controlled NS}} \\
& \multicolumn{3}{c}{$0.8\times$ training energy} & \multicolumn{3}{c}{$1.2\times$ training energy} \\
\cmidrule(lr){2-4}\cmidrule(lr){5-7}
Method & Energy CE & IC RelL2 & Div. RMS & Energy CE & IC RelL2 & Div. RMS \\
\midrule
Vanilla FM & \ensuremath{2.56\mathrm{e}{-1}} & 12.40\% & 0.0622 & \ensuremath{1.77\mathrm{e}{-1}} & 9.86\% & 0.0622 \\
Vanilla + projection & \ensuremath{2.10\mathrm{e}{-7}} & 3.92\% & 0.0558 & \ensuremath{2.38\mathrm{e}{-7}} & 4.37\% & 0.0683 \\
PCFM-Base / ECI-1 & \ensuremath{2.11\mathrm{e}{-7}} & 11.11\% & 0.1872 & \ensuremath{2.23\mathrm{e}{-7}} & 12.61\% & 0.2613 \\
PCFM-Relaxed (pre-proj.) & \ensuremath{2.33\mathrm{e}{-3}} & 11.37\% & 0.1879 & \ensuremath{1.04\mathrm{e}{-2}} & 14.01\% & 0.2923 \\
\quad + terminal projection & \ensuremath{2.20\mathrm{e}{-7}} & 11.38\% & 0.1881 & \ensuremath{1.99\mathrm{e}{-7}} & 13.98\% & 0.2913 \\
ECI-2 (fixed source) & \ensuremath{2.11\mathrm{e}{-7}} & 11.04\% & 0.1844 & \ensuremath{1.91\mathrm{e}{-7}} & 12.22\% & 0.2530 \\
ECI-2 (refreshed source) & \ensuremath{2.03\mathrm{e}{-7}} & 11.20\% & 0.1915 & \ensuremath{1.96\mathrm{e}{-7}} & 11.67\% & 0.2479 \\
Stop-grad. (199 calls) & \ensuremath{1.62\mathrm{e}{-1}} & 8.59\% & 0.0752 & \ensuremath{1.66\mathrm{e}{-1}} & 9.00\% & 0.0799 \\
Stop-grad. (99 calls) & \ensuremath{1.61\mathrm{e}{-1}} & 8.39\% & 0.0659 & \ensuremath{1.63\mathrm{e}{-1}} & 8.61\% & 0.0575 \\
Full-gradient guidance & \ensuremath{1.77\mathrm{e}{-1}} & 9.09\% & 0.1127 & \ensuremath{1.62\mathrm{e}{-1}} & 8.84\% & 0.0562 \\
CCFM ($n=1$) & \ensuremath{2.08\mathrm{e}{-7}} & 4.56\% & 0.0657 & \ensuremath{2.25\mathrm{e}{-7}} & 4.89\% & 0.1072 \\
CCFM ($n=2$) & \ensuremath{2.04\mathrm{e}{-7}} & 3.99\% & 0.0587 & \ensuremath{2.14\mathrm{e}{-7}} & 4.25\% & 0.1017 \\
MAFM (pre-proj.) & \ensuremath{2.07\mathrm{e}{-5}} & 2.87\% & 0.0399 & \ensuremath{1.44\mathrm{e}{-5}} & 3.45\% & 0.0685 \\
\quad + terminal projection & \ensuremath{2.10\mathrm{e}{-7}} & \textbf{2.87\%} & 0.0399 & \ensuremath{2.12\mathrm{e}{-7}} & \textbf{3.45\%} & 0.0685 \\
\bottomrule
\end{tabular*}
}
\end{table}

%% file: tables/doubled_energy.tex
\begin{table}[H]
\centering\small
\setlength{\tabcolsep}{2pt}
\caption{\rev{Comparisons at twice the training energy, with 128 paired
samples per system and 100 forward calls per sample. Bold marks the lowest
CE and the lowest relative $L_2$ among methods meeting the energy tolerance
(mean CE $\leq 10^{-5}$).}}
\label{tab:doubled-energy}
{\color{revisionink}
\begin{tabular*}{\linewidth}{@{\extracolsep{\fill}}lrrrrrr@{}}
\toprule
& \multicolumn{3}{c}{Euler} & \multicolumn{3}{c}{Controlled NS} \\
\cmidrule(lr){2-4}\cmidrule(lr){5-7}
Method & Energy CE & Rel.\ $L_2$ & Div.\ RMSE & Energy CE & Rel.\ $L_2$ & Div.\ RMSE \\
\midrule
PCFM-Relaxed + terminal projection & \ensuremath{2.20\mathrm{e}{-7}} & 37.92\% & 0.805 & \ensuremath{1.96\mathrm{e}{-7}} & 25.20\% & 0.745 \\
ECI-2 (refreshed source) & \textbf{\boldmath\ensuremath{1.68\mathrm{e}{-7}}} & 35.03\% & 0.785 & \textbf{\boldmath\ensuremath{1.74\mathrm{e}{-7}}} & 25.56\% & 0.731 \\
Full-gradient guidance & \ensuremath{5.00\mathrm{e}{-1}} & 30.72\% & 0.061 & \ensuremath{4.92\mathrm{e}{-1}} & 30.53\% & 0.066 \\
CCFM ($n=2$) & \ensuremath{2.20\mathrm{e}{-7}} & 22.96\% & 0.655 & \ensuremath{2.12\mathrm{e}{-7}} & 19.25\% & 0.646 \\
MAFM + terminal projection & \ensuremath{1.86\mathrm{e}{-7}} & \textbf{9.85\%} & 0.210 & \ensuremath{1.94\mathrm{e}{-7}} & \textbf{10.01\%} & 0.222 \\
\bottomrule
\end{tabular*}
}
\end{table}

%% file: tables/additional_inverse_baselines.tex
\begin{table}[H]
\centering\small
\setlength{\tabcolsep}{2pt}
\caption{\rev{Supplemental comparisons on five PDE benchmarks. CE is the supplied-constraint error. Rel.\ $L_2$ is the own-initial-condition rollout error for time-dependent PDEs and the error against an independent solve from the generated source for Poisson. Div.\ RMSE measures deviation from zero divergence. Bold denotes the lowest value in each column at the displayed precision.}}
\label{tab:additional-inverse-baselines}
{\color{revisionink}
\begin{tabular*}{\linewidth}{@{\extracolsep{\fill}}lrrrrrr@{}}
\toprule
& \multicolumn{3}{c}{Euler} & \multicolumn{3}{c}{Controlled NS} \\
\cmidrule(lr){2-4}\cmidrule(lr){5-7}
Method & Energy CE & Rel.\ $L_2$ & Div.\ RMSE & Energy CE & Rel.\ $L_2$ & Div.\ RMSE \\
\midrule
CCFM ($n=2$) & \textbf{\boldmath\ensuremath{2.33\mathrm{e}{-7}}} & 3.68\% & 0.06499 & \textbf{\boldmath\ensuremath{2.35\mathrm{e}{-7}}} & 3.48\% & 0.06402 \\
D-Flow & 0.00751 & 3.54\% & 0.06256 & 0.00748 & 3.27\% & 0.05994 \\
PnP-Flow & 0.0581 & 11.33\% & 0.20279 & 0.0554 & 10.25\% & 0.18868 \\
LDF (198 NFE) & \ensuremath{5.85\mathrm{e}{-4}} & 7.13\% & 0.16506 & \ensuremath{5.35\mathrm{e}{-4}} & 5.86\% & 0.15182 \\
MAFM & \ensuremath{6.84\mathrm{e}{-6}} & \textbf{2.94\%} & \textbf{0.04829} & \ensuremath{6.89\mathrm{e}{-6}} & \textbf{2.76\%} & \textbf{0.04736} \\
MAFM + terminal projection & \ensuremath{2.37\mathrm{e}{-7}} & \textbf{2.94\%} & \textbf{0.04829} & \ensuremath{2.43\mathrm{e}{-7}} & \textbf{2.76\%} & \textbf{0.04736} \\
\bottomrule
\end{tabular*}
\par\medskip
\begin{tabular*}{\linewidth}{@{\extracolsep{\fill}}lrrrrrrrr@{}}
\toprule
& \multicolumn{2}{c}{Burgers} & \multicolumn{2}{c}{Reaction--diffusion}
& \multicolumn{2}{c}{Poisson (64)} & \multicolumn{2}{c}{Poisson (256)} \\
\cmidrule(lr){2-3}\cmidrule(lr){4-5}\cmidrule(lr){6-7}\cmidrule(lr){8-9}
Method & Mass CE & Rel.\ $L_2$ & Mass CE & Rel.\ $L_2$
& Obs.\ CE & Rel.\ $L_2$ & Obs.\ CE & Rel.\ $L_2$ \\
\midrule
CCFM ($n=2$) & \ensuremath{5.39\mathrm{e}{-8}} & 5.73\% & \textbf{\boldmath\ensuremath{5.96\mathrm{e}{-8}}} & 6.83\% & \textbf{0} & 22.78\% & \textbf{0} & 46.47\% \\
D-Flow & 0.995 & 39.01\% & 0.0317 & 3.45\% & 0.555 & \textbf{1.03\%} & 0.567 & 1.04\% \\
PnP-Flow & 0.00149 & \textbf{0.71\%} & 0.00221 & \textbf{0.98\%} & 0.538 & 1.22\% & 0.234 & \textbf{1.00\%} \\
LDF (198 NFE) & \ensuremath{7.22\mathrm{e}{-3}} & 4.54\% & \ensuremath{2.69\mathrm{e}{-3}} & 21.81\% & \ensuremath{3.84\mathrm{e}{-2}} & 14.96\% & \ensuremath{6.65\mathrm{e}{-2}} & 17.53\% \\
MAFM & \textbf{\boldmath\ensuremath{4.22\mathrm{e}{-8}}} & 0.965\% & \ensuremath{1.13\mathrm{e}{-7}} & 2.74\% & \ensuremath{1.72\mathrm{e}{-7}} & 1.70\% & \ensuremath{1.75\mathrm{e}{-7}} & 2.25\% \\
MAFM + terminal projection & \ensuremath{5.32\mathrm{e}{-8}} & 0.965\% & \ensuremath{6.77\mathrm{e}{-8}} & 2.74\% & \textbf{0} & 1.70\% & \textbf{0} & 2.25\% \\
\bottomrule
\end{tabular*}
}
\end{table}